\documentclass[11pt]{article}

\usepackage{acl}

\usepackage{times}
\usepackage{latexsym}

\usepackage[T1]{fontenc}

\usepackage[utf8]{inputenc}

\usepackage{microtype}

\usepackage{inconsolata}

\usepackage{graphicx}

\newtheorem{problem}{Problem}
\newtheorem{task}{Task}
\usepackage{amsmath}
\usepackage{multirow}
\usepackage{multicol}
\usepackage{booktabs}
\usepackage{amssymb}
\usepackage{array}
\usepackage{adjustbox}

\usepackage{tabularx}
\usepackage{xurl}
\usepackage{hyperref}
\usepackage[skins]{tcolorbox}
\tcbuselibrary{breakable}
\usepackage{float}
\usepackage{pifont}
\usepackage[table,xcdraw]{xcolor}
\usepackage[dvipsnames]{xcolor}
\usepackage{threeparttable}
\usepackage{listings}

\newcommand{\blackcircnum}[1]{\ding{\numexpr201+#1\relax}}

\newtcolorbox[auto counter, number within=section]{prompt}[3][]{%
  enhanced,
  breakable,
  colback=#2!5!white,
  colframe=#2!75!black,
  title=\textbf{Box \thetcbcounter: #3},
  fontupper=\normalsize\fontfamily{cmss}\selectfont,
  #1
}

\newtcolorbox{questionbox}{
enhanced,
boxrule=0pt,frame hidden,
borderline west={4pt}{0pt}{Emerald!75!black},
colback=Emerald!9!white,
sharp corners
}

\newtcolorbox{taskbox}{
enhanced,
boxrule=0pt,frame hidden,
borderline west={4pt}{0pt}{SkyBlue!55!black},
colback=SkyBlue!9!white,
sharp corners
}

\lstdefinestyle{myjson}{
  basicstyle=\ttfamily\footnotesize,
  backgroundcolor=\color{gray!5},
  frame=single,
  breaklines=true,
  columns=fullflexible
}

\lstdefinestyle{mytext}{
  basicstyle=\ttfamily\footnotesize,
  backgroundcolor=\color{gray!3},
  frame=single,
  breaklines=true,
  columns=fullflexible
}

\title{\textsc{ScriptMind}: Crime Script Inference and Cognitive Evaluation for LLM-based Social Engineering Scam Detection System}

\author{
  \textbf{Heedou Kim\textsuperscript{1,2}}, 
  \textbf{Changsik Kim\textsuperscript{2}}, 
  \textbf{Sanghwa Shin\textsuperscript{3,*}}, 
  \textbf{Jaewoo Kang\textsuperscript{1,*}} \\
  \textsuperscript{1}Korea University, Seoul, Republic of Korea, \\
  \textsuperscript{2}Korean National Police Agency, Seoul, Republic of Korea, \\
  \textsuperscript{3}Inje University, Gimhae, Republic of Korea \\
    \textsuperscript{*}Co-corresponding authors.
}

\begin{document}
\maketitle
\begin{abstract}
Social engineering scams increasingly employ personalized, multi-turn deception, exposing the limits of traditional detection methods. While Large Language Models (LLMs) show promise in identifying deception, their cognitive assistance potential remains underexplored. We propose \textbf{\textsc{ScriptMind}}, an integrated framework for LLM-based scam detection that bridges automated reasoning and human cognition. It comprises three components: the Crime Script Inference Task (\textbf{CSIT}) for scam reasoning, the Crime Script–Aware Inference Dataset (\textbf{CSID}) for fine-tuning small LLMs, and the Cognitive Simulation-based Evaluation of Social Engineering Defense (\textbf{CSED}) for assessing real-time cognitive impact. Using 571 Korean phone scam cases, we built 22,712 structured scammer-sequence training instances. Experimental results show that the 11B small LLM fine-tuned with \textbf{\textsc{ScriptMind}} outperformed GPT-4o by 13\%, achieving superior performance over commercial models in detection accuracy, false-positive reduction, scammer utterance prediction, and rationale quality. Moreover, in phone scam simulation experiments, it significantly enhanced and sustained users’ suspicion levels, improving their cognitive awareness of scams. \textbf{\textsc{ScriptMind}} represents a step toward human-centered, cognitively adaptive LLMs for scam defense.\\
\textbf{\mbox{Data \& Code:}} \href{https://anonymous.4open.science/r/ScriptMind-8D48}{anonymous/ScriptMind} \\
\end{abstract}

\section{Introduction}
Preventing social engineering scams is essential for financial security, psychological protection, and societal trust. Online scams have grown sophisticated, demanding more adaptive defenses. In this context, Large Language Models (LLMs) have emerged as interpretable, cognitively assistive tools capable of detecting deception and enhancing user awareness \cite{lim2025explicate}, proving effective in brand impersonation, fake webpage detection, and phone scam monitoring as real-time defense systems against evolving social engineering scams \cite{koide2024chatspamdetector, lee2024multimodal, shen2025warned}.

\begin{figure}[t]
    \centering
    \includegraphics[width=1\columnwidth]{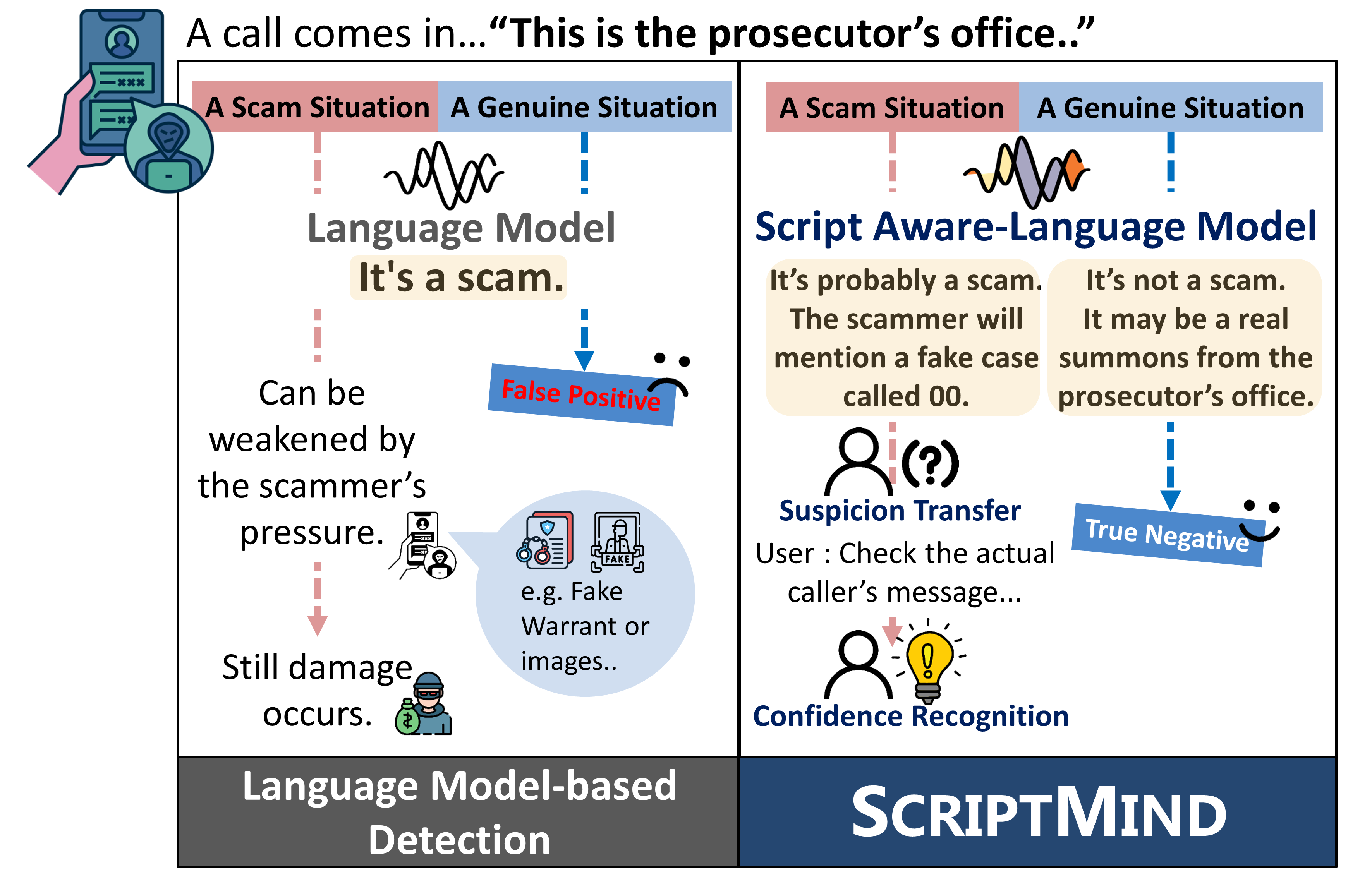} 
    \caption{Scam alerts provide accurate detection and explanations but can be neutralized by new tactics. \textbf{\textsc{ScriptMind}} overcomes these limits through a crime script inference and simulation-based evaluation, enabling cognitively effective scam defense.}
    \label{fig:fig1}
\end{figure}

However, social engineering scams have become increasingly sophisticated, using psychological tactics that neutralize suspicion. Scammers exploit user's anxiety and shifting doubt through multiple strategies, leading to psychological submission~\cite{han2024voice, wang2021social}. Thus, detection must move beyond scam identification to consider how and when warnings are cognitively delivered. As shown in Figure~\ref{fig:fig1}, alerts themselves can be manipulated. For example, scammers may counter a “fake prosecutor” warning by invoking legal pressure by presenting fabricated court documents. False positives in benign interactions can also erode system trust. Therefore, effective defense requires modeling the temporal dynamics of suspicion and designing adaptive cognitive assistants that respond to users’ evolving mental states.

Most existing LLM-based social engineering scam detection studies focus primarily on identifying deceptive content, without adequately reflecting how users’ cognitive and behavioral responses evolve in scam situations. In addition, little attention has been paid to how warning alerts influence users’ suspicion levels or what types of feedback effectively enhance scam awareness \cite{kumarage2025personalized}. These limitations have become increasingly critical as social engineering scams grow more personalized, weakening the distinction between LLM-based methods and traditional approaches such as blacklists, phishing campaigns, or conventional automated detection. Therefore, it is essential to validate the cognitive capability of LLMs and develop a framework that dynamically strengthens user suspicion and systematically evaluates its effectiveness in real-world scam scenarios.

We introduce \textbf{\textsc{ScriptMind}}, a framework designed to integrate LLM-based inference for effective social engineering scam detection with user-centered evaluation of acceptability. \textbf{\textsc{ScriptMind}} supports users cognitively during real-time interactions with scammers by introducing a novel detection task that predicts the scammer’s crime script. Through this process, it observes meaningful changes in the user’s level of suspicion at each conversational stage and evaluates its effectiveness.

Core three components of \textbf{\textsc{ScriptMind}} are: the \textbf{C}rime \textbf{S}cript \textbf{I}nference \textbf{T}ask (\textbf{CSIT}), which models reasoning over scam scripts; the \textbf{C}rime \textbf{S}cript–Aware \textbf{I}nference \textbf{D}ataset (\textbf{CSID}), designed to support efficient and secure fine-tuning of LLMs; and the \textbf{C}ognitive \textbf{S}imulation-based \textbf{E}valuation of Social Engineering \textbf{D}efense (\textbf{CSED}) for evaluating LLM-driven scam detection from a user acceptance perspective. To the best of our knowledge, this is the first approach that unifies scam detection with cognitive effectiveness evaluation.

The \textbf{CSIT} and \textbf{CSID} was designed to model how an LLM assists users through the cognitive shift from suspicion to conviction during scam interactions. Using crime script analysis, we extracted scammer behavior patterns and formulated a task enabling the model to infer and explain scam intent at each dialogue stage. From publicly available phone scam cases in Korea, we built 22,712 crime script prediction instances for LLM training, including a benign dataset that contains scenarios of legitimate police summons. We then verified the statistical significance of extracted patterns and evaluated the fine-tuning performance gains.

To evaluate user cognitive effects through \textbf{CSED}, we conducted a phone scam simulation experiment in which participants were instructed, \textit{“The caller may be a real prosecutor or a scammer; listen carefully and decide.”} Participants were sequentially presented with 40 structured utterances representing key criminal intents of a scammer, delivered in both audio and text. The experiment consisted of three conditions: no AI intervention, a single AI warning, and LLM-based stepwise explanatory assistance. Participants’ suspicion levels at each major stage of the dialogue were measured using a Likert scale to assess how different forms of intervention influenced suspicion escalation.

Experimental results show that \textbf{\textsc{ScriptMind}}'s finetuned sLLM achieved notable gains in scam detection, utterance prediction, and intent explanation over the baseline, outperforming commercial zero-shot models by about 13\%. Moreover, leveraging next-utterance prediction to guide users’ cognitive shift from suspicion to conviction, the \textbf{\textsc{ScriptMind}} significantly raised participants’ suspicion levels compared to single-warning and no-intervention conditions. These results demonstrate \textbf{\textsc{ScriptMind}}’s effectiveness in enhancing user engagement and advancing LLM-based defenses toward practical crime prevention. Building on this, we present a UI/UX-oriented design for detection system(Appendix~\ref{UI}). Contributions are:

\begin{enumerate}
    \item \textbf{\textsc{ScriptMind}}: We propose the first framework that unifies LLMs’ cognitive assistance in elevating user suspicion during real social engineering scams, together with a simulation-based evaluation method for user acceptance.
    \item We constructed a \textbf{Crime Script Inference Task} and 22,712 training instances for LLM.
    \item We empirically analyze the effectiveness of fine-tuned smaller LLMs operating in resource-constrained environments.
    \item Through a phone scam simulation, \textbf{\textsc{ScriptMind}} demonstrated a significant increase in users’ suspicion compared to control groups.
\end{enumerate}

\begin{figure*}[h!]
    \centering
    \includegraphics[width=0.9\textwidth]{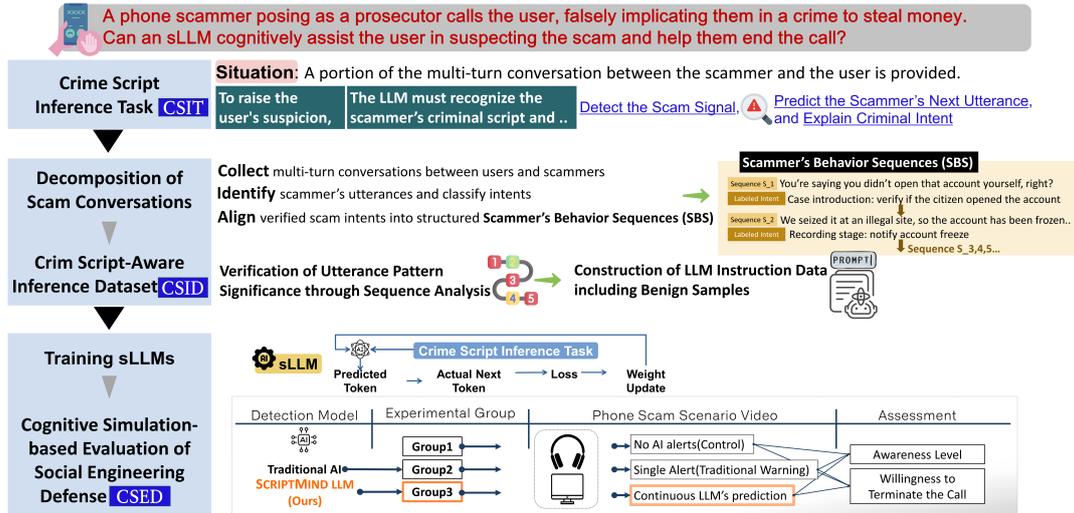} 
    \caption{The uniqueness of \textbf{\textsc{ScriptMind}} lies in modeling and evaluating tasks that elicit and reinforce users’ suspicion throughout scam interactions. Unlike prior studies that evaluate models primarily based on accuracy\cite{koide2024chatspamdetector, lee2024multimodal, shen2025warned}, we design a framework that trains LLMs to predict scammers’ next scripted actions and assess whether such predictions meaningfully enhance user suspicion in realistic scam contexts.}
    \label{fig:overview}
\end{figure*}

\begin{table*}[h!]
\centering
\renewcommand{\arraystretch}{0.8}
\resizebox{\textwidth}{!}{
\begin{tabular}{lcp{5cm}p{5.5cm}p{7cm}}  
\toprule
\textbf{Category} &
\textbf{Original Dataset} &
\textbf{Initial Dataset} &
\textbf{Scammer's Behavior Sequences \footnotesize(ours)} &
\textbf{Crime Script-Aware Instruction Dataset \footnotesize(ours)} \\
\midrule
\textbf{Purpose} &
Scam Conversations Collection&
Identifying Scammer's Intention in Partial Conversation &
Statistical Validation of Scammer's Utterance Pattern &
Scam Prediction, Utterance Prediction, Rationale Explanation using LLMs \\
\midrule
\textbf{Instance Type} &
$\displaystyle D_o = \{ \mathcal{C} \}$ &
$\displaystyle D_I = \{ (U_S, Y_{intent}) \}$ &
$\displaystyle D_{\text{SBS}} = \{ (U^{\text{scm}}, Y_{intent}) \}$ &
$\displaystyle D_{\text{CSID}} = \{ (U_S, Y_a, U_{t+1}^{\text{scm}}, Y_{intent}) \}$ \\
\midrule
\textbf{Scam Cases} & 571 & 571 & 571 & 571 \\
\textbf{Benign Conversations} & -- & -- & -- & 11,356 \\
\midrule
\textbf{All Data Instances} & 571 & 23,771 & 571 & 22,712 \\
\bottomrule
\end{tabular}}
\caption{Summary of Scam Scenario Data Used for the \textbf{C}rime \textbf{S}cript–Aware \textbf{I}nference \textbf{D}ataset (\textbf{CSID}) Construction.}
\label{datasummary}
\end{table*}

\section{Related Work}

\subsection{Evolution of Social Engineering Scams}
Online scam exploits human psychological vulnerabilities through social engineering to steal sensitive information \cite{wang2021social}. With advances in LLMs and generative AI, such scams have evolved into large-scale, organized, and multi-channel attacks \cite{INTERPOL2024, apwg2024q4, FBI2023, proofpoint2024state, jorij2024scams}. Scammers now use SMS, calls, social media, and deepfakes in multi-turn conversations, impersonating acquaintances, recruiters, or officials to build trust and extract data \cite{tsinganos2018towards, zheng2019session, Reuters2023, kumarage2025personalized, ai2024defending, fss_voicephishing}. Global damages include a \$600K deepfake scam in China, \$1.3B in U.S. elder fraud, and ₩190B in South Korean voice phishing cases \cite{Reuters2023, fss_voicephishing, FBI2023}.

\subsection{LLM-based Scam Detection}
Language model–based scam detection serves as the core engine of modern anti-phishing systems \cite{cao2025phishagent, koide2024chatspamdetector, koide2023detecting, lee2024korsmishing, yu2024korean}. Traditional classifiers lack explainability, whereas recent dialog-based frameworks enhance interpretability through scenario-driven detection grounded in social engineering contexts \cite{lee2024korsmishing, koide2024chatspamdetector, lim2025explicate}. Such detection has also expanded into multimodal domains, including fake website analysis, brand impersonation detection, and real-time conversational alerts, demonstrating strong potential for AI-driven warnings. \cite{lee2024multimodal, kulkarni2025ml, shen2025warned}.

Still, prior studies overlook user cognition and behavioral responses. This approach fails to address overconfidence and alert fatigue often seen in traditional defenses \cite{redmiles2016think, wang2016overconfidence, vishwanath2018suspicion, merete2008implementation, wang2021incentive}.

\section{Method}

Our core idea for preventing social engineering scams is to use LLMs to strengthen users’ scam awareness and guide their decision to end the conversation. While detection accuracy is important, it is ultimately the user who decides to terminate the interaction, and scammers exploit psychological manipulation to stop them from doing so. We propose that enabling users to anticipate scammers’ next actions, much like scammers anticipate users’ vulnerabilities, can turn suspicion into conviction. To realize this, as illustrated in Figure~\ref{fig:overview}, the \textbf{CSIT} models scammers’ behavioral sequences to train LLMs to predict their next moves. The fine-tuned models, based on the \textbf{CSID}, are then evaluated through scam simulations (\textbf{CSED}). This section details the construction of these three components.

\subsection{Task Formulation}
\begin{problem}[Enhancing Users’ Scam Awareness]
\label{problem:sai_task}
The purpose of the social engineering scam prevention system is to build a cognitive assistance–based detection service that helps users recognize and confirm fraudulent intent during real-time conversations without third-party intervention. Given a conversation context $\mathcal{C}$ within an unknown stage of a scam scenario $\mathcal{S}$, the model generates supportive inference that foster user suspicion and confidence to safely terminate the conversation. 
\end{problem}

\begin{task}[Crime Script Inference Task, CSIT]
\label{task:crime_script_inference}
Given an input \textbf{prompt} $\mathbf{X}$ containing a conversation $\mathcal{C}$ under a potential scam scenario $\mathcal{S}$, the model $F$ jointly performs scam detection, next-utterance prediction, and intent inference to simulate cognitive reasoning in real-time scam interactions. The task is defined such that {\small $F(\mathbf{X}) = \{y_a, U_{t+1}, y_i\}$, where $\mathbf{X} = \{\mathcal{S}, \mathcal{C}\}$, $y_a \in [0,1]$, $U_{t+1} \in \mathcal{U}$, and $y_i \in \mathcal{Y}_{\text{intent}}$}.

\end{task}

\subsection{Dataset Construction}

\subsubsection{Decomposition of Scam Conversations}
\paragraph{Scam Conversations} The original data were collected from the publicly available Law\&Order Benchmark~\cite{lawandorder2025}. The dataset was constructed based on voice phishing call records released by the Korean National Police Agency, comprising conversations in which scammers impersonate prosecutors to fabricate criminal cases and exert financial pressure on victims. Such impersonation and coercive tactics, using false legal threats, are typical scam strategies observed across multiple countries~\cite{FBI2023, INTERPOL2024}. We utilized a total of 571 cases and 48,229 utterances included in the the dataset(see Appendix~\ref{sec:appendix}).

\paragraph{Scammer's Behavior Sequences} To implement the \textbf{CSIT} using the given scam conversations, we first separated the scammer’s utterances from each dialogue and labeled those that explicitly conveyed fraudulent intent. Through this process, utterances that repeatedly appeared across confirmed scam cases were organized into structured data referred to as Scammer's Behavior Sequences(SBS). This sequence-based analysis is theoretically grounded in Crime Script Analysis, which models social engineering as sequential behavioral scripts~\cite{cornish1994crimes, hutchings2015crime, loggen2022unraveling, choi2017voice, lwin2023supporting}, and the MITRE ATT\&CK framework, which systematically categorizes phishing techniques~\cite{strom2018mitre, shin2022focusing, abo2024application}.
As shown in Table~\ref{datasummary}, the initial dataset contained 23,771 scammer utterances, each potentially associated with multiple intents. We segmented the dialogues and mapped verified intents to individual utterances, normalizing the data into a single-utterance–single-intent format (Appendix~\ref{SBS}, ~\ref{Scenario_Details}).

\subsubsection{Crime Script-Aware Inference Dataset}

\paragraph{Statistical Validation of Sequences} To verify that the classified utterance sequences reflected consistent criminal patterns, we performed statistical validation to distinguish scripted behaviors from improvised statements. Weak sequence associations, even with expert-labeled intents, can undermine the reliability of key behavioral patterns. To address this, we applied the Standardized Residual (SR) method from Behavior Sequence Analysis~\cite{everitt2010cambridge}. The SR score quantifies the normalized difference between the observed and expected frequencies of intent transitions across utterances. A higher SR indicates more consistent and scripted scammer behavior, aligning with criminological perspectives that interpret repetitive behaviors as indicators of intentional or patterned actions~\cite{cornish1994crimes}. Details of the SR analysis are provided in Appendix~\ref{SR_analysis}.

\paragraph{Crime Script-Aware Inference Dataset} Using the Scammer’s Behavior Sequences, we constructed the \textbf{C}rime \textbf{S}cript-aware \textbf{I}nference \textbf{D}ataset (\textbf{CSID}), designed to enable the LLM to perform tasks based on partial conversational(Appendix~\ref{CSID}). In addition, an equal number of benign instances were added under the scenario of “a legitimate police officer issuing a summons” (Table~\ref{benign}). For scam instances, original dialogues were segmented into input–output pairs, where only preceding utterances were provided as input. Consequently, it supports (1) scam detection, (2) utterance prediction, and (3) intent explanation.

\subsection{Training Smaller Large Language Model}
We trained an open-source \emph{compact LLM (cLLM)} to support closed-network deployment and lightweight, privacy-preserving operation in restricted environments. We hypothesized that additional training on expert-labeled scammer interaction sequences is necessary to compensate for the model’s limited domain-specific prior knowledge and to better align its reasoning with real-world scam scenarios.

We evaluated models at three capacity ranges: \textbf{1--2B}, \textbf{7--11B}, and \textbf{large commercial models}, including those specialized for Korean. The open-source cLLMs were fine-tuned for 5 epochs using the Paged AdamW optimizer (learning rate = $1\mathrm{e}{-4}$). Parameter-efficient adaptation was performed using QLoRA~\cite{dettmers2023qlora}, with low-rank adapters applied to the attention and feed-forward layers. Training was conducted on two NVIDIA A100 80\,GB GPUs, requiring approximately 30 hours (Appendix~\ref{modelselction}).

\subsection{Cognitive Simulation-based Evaluation}
\paragraph{Aim and Hypotheses} We aim to examine whether \textbf{\textsc{ScriptMind}} can serve as a cognitive assistant that supports users’ real-time judgment during scam. We hypothesize that \textit{real-time LLM warnings and explanations enhance users’ suspicion levels.} To test this, a five-stage conversational script based on phone scam cases was designed, and participants reported their suspicion levels in real time while listening to both audio and text. 

\paragraph{Experimental Stimuli and Procedure} We used \textit{prosecutor impersonation scam scenario} from the \textbf{CSID} as the experimental stimulus, and the structured scam script is provided in Appendix~\ref{Scenario_Details}. 98 participants were recruited and evenly assigned across conditions. We used repeated-measures ANOVA and t-tests to assess the impact of AI intervention. All procedures were IRB approved and details regarding the purpose and scope of the experiment (Appendix~\ref{aim_scope}), the design of experimental stimuli and procedures (Appendix~\ref{design}), the questionnaire items (Appendix~\ref{questions}), the statistical analysis methods used for interpretation (Appendix~\ref{statistics_method}), and the ethical review and approval for human research (Appendix~\ref{Ethics}) are all provided.

\begin{taskbox}
\label{task:scriptmind_conditions}
\textbf{\textcolor{Emerald!55!black}{\small \textsc{Experimental Conditions}}} \\
\textit{Participants were told each call might be legitimate or fraudulent, prompting cognitive judgment. Conditions: (1) \textbf{a control group} with no alerts, (2) \textbf{a single-warning group} with one alert during the financial information stage, (3) \textbf{a \textsc{ScriptMind}'s LLM group} providing real-time predicted utterances at each scam stage.}
\end{taskbox}

\section{Experiment Results}

\begin{table*}[h!]
\centering
\scalebox{0.8}{
\begin{tabular}{|l|l|ccc|c|c|}
\hline
\multicolumn{1}{|c|}{\multirow{2}{*}{\textbf{Model}}} & 
\multicolumn{1}{c|}{\multirow{2}{*}{\textbf{Method}}} & 
\multicolumn{3}{c|}{\textbf{Scam Detection}} & 
\multicolumn{1}{c|}{\textbf{Next Utterance}} & 
\multicolumn{1}{c|}{\textbf{Intent Inference}} \\ \cline{3-7}
\multicolumn{1}{|c|}{} & 
\multicolumn{1}{c|}{} & 
\textbf{ACC} & \textbf{F1} & \textbf{FP/FN} & 
\textbf{LLM-as-a-Judge} & 
\textbf{LLM-as-a-Judge} \\ \hline

\texttt{Llama-3.2-1B-Instruct} & ZS & 0.55 & 0.56 & 0.24/0.21 & 0.04 & 0.17 \\ 
                               & \textbf{\textsc{ScriptMind}}-FT & 0.91 & 0.92 & 0.06/0.02 & 0.39 & 0.57 \\ \hline

\texttt{EXAONE-3.5-2.4B-Instruct} & ZS & 0.58 & 0.65 & 0.31/0.11 & 0.30 & 0.51 \\ 
                                 & \textbf{\textsc{ScriptMind}}-FT & \textbf{\underline{0.94}} & \textbf{\underline{0.94}} & 0.06/\textbf{\underline{0.01}} & \textbf{\underline{0.53}} & 0.73 \\ \hline

\texttt{Midm-2.0-Mini-Instruct} & ZS & 0.48 & 0.44 & 0.23/0.29 & 0.11 & 0.29 \\ 
                               & \textbf{\textsc{ScriptMind}}-FT & 0.74 & 0.67 & 0.03/0.23 & 0.33 & 0.53 \\ \hline

\texttt{Llama-3.1-8B-Instruct} & ZS & 0.50 & 0.67 & \cellcolor{gray!20}0.50/\textbf{\underline{0.00}} & 0.19 & 0.54 \\ 
                              & \textbf{\textsc{ScriptMind}}-FT & 0.80 & 0.76 & 0.01/0.19 & 0.41 & 0.54 \\ \hline

\texttt{SOLAR-10.7B-Instruct} & ZS & 0.64 & 0.72 & 0.33/0.03 & 0.16 & 0.44 \\ 
                             & \textbf{\textsc{ScriptMind}}-FT & 0.85 & 0.82 & \cellcolor{gray!20}\textbf{\underline{0.00}}/0.15 & 0.44 & 0.50 \\ \hline

\texttt{EEVE-Korean-Instruct-10.8B} & ZS & 0.71 & 0.74 & 0.21/0.09 & 0.42 & 0.52 \\ 
                                   & \textbf{\textsc{ScriptMind}}-FT & \cellcolor{gray!20}\textbf{\underline{0.98}} & \cellcolor{gray!20}\textbf{\underline{0.98}} & \textbf{\underline{0.01}}/\textbf{\underline{0.01}} & \cellcolor{gray!20}\textbf{\underline{0.68}} & \cellcolor{gray!20}\textbf{\underline{0.80}} \\ \hline

\texttt{EXAONE-3.0-7.8B-Instruct} & ZS & 0.64 & 0.72 & 0.32/0.04 & 0.26 & 0.63 \\ \hline

\texttt{Midm-2.0-Base-Instruct} & ZS & 0.55 & 0.6 & 0.28/0.17 & 0.22 & 0.57 \\ 
                               & \textbf{\textsc{ScriptMind}}-FT & 0.73 & 0.64 & \cellcolor{gray!20}\textbf{\underline{0.00}}/0.26 & 0.33 & 0.5 \\ \hline

\texttt{chatgpt-4o-latest} & ZS & 0.90 & 0.91 & \cellcolor{gray!20}0.1/\textbf{\underline{0.00}} & 0.45 & \textbf{\underline{0.78}} \\ \hline
\texttt{gemini-2.0-flash} & ZS & 0.70 & 0.77 & \cellcolor{gray!20}0.29/\textbf{\underline{0.00}} & 0.39 & 0.77 \\ \hline
\texttt{claude-3-5-haiku} & ZS & 0.67 & 0.75 & \cellcolor{gray!20}0.33/\textbf{\underline{0.00}} & 0.31 & 0.73 \\ \hline

\end{tabular}
}
\caption{Evaluation results of LLMs on our tasks.  
\textbf{ZS} indicates zero-shot and \textbf{\textbf{\textsc{ScriptMind}}-FT} refers to finetuning.}
\label{tab:scriptmind_results}
\end{table*}

\begin{table*}[h]
\renewcommand{\arraystretch}{0.9}
\centering
\label{tab:stage_group_stats}
\footnotesize
\begin{tabular}{p{6cm} p{1.9cm} p{2cm} p{1.5cm} p{0.7cm} p{0.7cm}}
\hline
\textbf{Stage} & \textbf{\textsc{ScriptMind}} & \textbf{Single\_Warning} & \textbf{Control} & \textbf{F} & \textbf{P} \\
\hline
1.Introduction of a fake case & \textbf{5.63}±1.69 & 5.40±1.92 & 5.07±2.07 & 0.67 & .512 \\
\hline
2.Explanation of alleged criminal involvement & \textbf{5.60}±1.65 & 4.77±2.08 & 4.80±1.90 & 1.88 & .159 \\
\hline
3.Setup of a recorded investigation & \textbf{4.63}±1.99 & 3.87±2.21 & 4.13±1.96 & 1.07 & .346 \\
\hline
4.Request for financial information & \cellcolor{gray!20}\textbf{\underline{6.27}}±1.60 & 6.00±1.49 & 5.23±1.72 & 3.36 & \cellcolor{gray!20}\textbf{\underline{.039}} \\
\hline
5.Notice of summons for investigation & \cellcolor{gray!20}\textbf{\underline{5.73}}±2.05 & 5.63±1.69 & 4.43±2.25 & 3.88 & \cellcolor{gray!20}\textbf{\underline{.024}} \\
\hline
\end{tabular}
\caption{Suspicion scores by stage and group. We conducted stage-wise ANOVA. The \textbf{\textsc{ScriptMind}} showed significantly higher suspicion at 4\textasciitilde5 (p = .039, .024), indicating statistical significance at the p < .05 level. This supports our research hypothesis that \textit{the LLM warning increases the level of suspicion more effectively than in other groups}. Detailed analysis of experimental results is presented in Appendix~\ref{RQ3}.}
\label{tab:suspicion}
\end{table*}

\noindent \textbf{Metrics} Scam detection was evaluated using Accuracy, F1, FP, FN, while utterance prediction and intent inference were assessed via the LLM-as-a-Judge~\cite{chiang2023can,lee2026evaluating}.
Strong correlation with expert evaluations confirmed the reliability of the automatic assessment. To validate the reliability of the LLM-as-a-Judge evaluation, we measured its agreement with human judgment on a randomly sampled subset of the test data. Specifically, two domain experts independently rated 200 instances each for the zeroshot and finetuned settings, following the same evaluation criteria as the LLM. Pearson correlation analysis ($p < 0.05$) showed strong alignment between human ratings and automatic scores, supporting the validity of the proposed evaluation protocol(Appendix~\ref{Finetuning_Analysis}).

\noindent \textbf{Fine-tuned Model Performance} \textbf{\textsc{ScriptMind}} consistently outperformed commercial models, demonstrating its effectiveness in enhancing scam awareness. As shown in Table~\ref{tab:scriptmind_results}, EEVE-Korean-10.8B achieved the best overall performance(Scam Detection 0.98, Next Utterance 0.68, Intent Inference 0.80), exceeding GPT-4o by 13\%. On average, fine-tuned small models showed a 51\% improvement over zero-shot (Table~\ref{tab:zs_results}). It also reduced false positives and improved explanatory quality: commercial models averaged FP 0.24, while fine-tuning achieved 0.02.

\begin{figure}[h]
\centering
\includegraphics[width=0.48\textwidth]{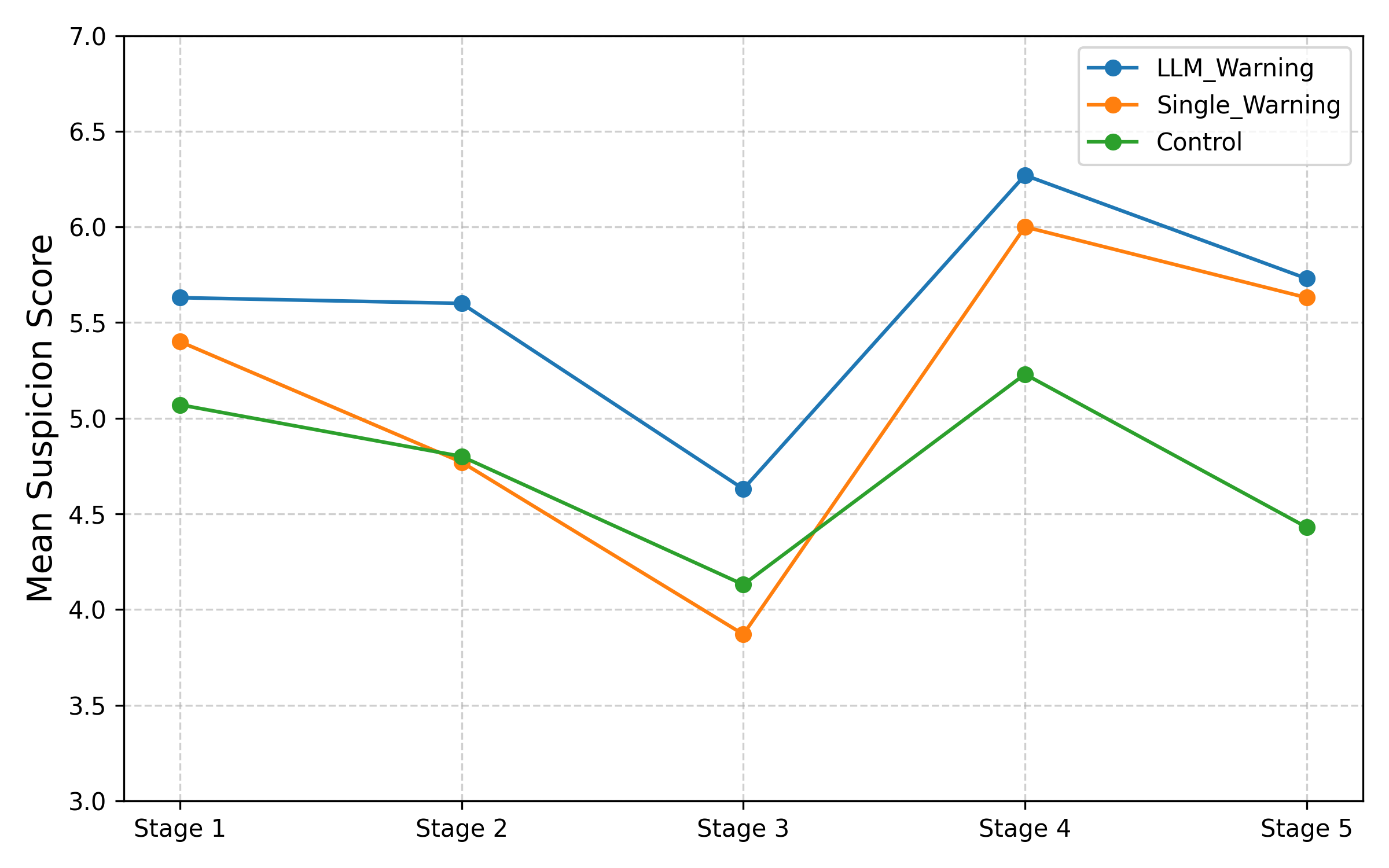}
\caption{Changes in Suspicion Levels by Script Stage.}\label{graph_script}
\end{figure}

\noindent \textbf{Cognitive Effect of \textbf{\textsc{ScriptMind}}} 
As shown in Figure~\ref{graph_script} and Table~\ref{tab:suspicion}, participants’ suspicion levels across the five-stage scam scenario demonstrated that \textbf{\textsc{ScriptMind}} next-utterance prediction warnings achieved the strongest cognitive resilience. Since real scam calls could not be ethically replicated, participants were informed in advance of potential scam, resulting in high initial suspicion. To control for this, statistical analyses were conducted to isolate the true effects of LLM intervention(Appendix~\ref{researchquestions}). First, we found that suspicion varied significantly across stages, whereas anxiety and relevance remained stable, confirming that suspicion can serve as a key indicator of cognitive resilience(Table~\ref{tab:stage_means}). It declined through Stages 1\textasciitilde3 but rose sharply at Stage 4, where monetary information was requested. Second, no significant difference was found between the single-warning and control groups (Table~\ref{tab:awareness_diff}), suggesting that one-time alerts yield only temporary awareness. However, a significant stage–group interaction (Table~\ref{tab:anova_results}), together with the stage-wise one-way ANOVA results, showed that \textbf{\textsc{ScriptMind}} maintained higher suspicion compared to all other groups, particularly at Stages 4\textasciitilde5 (Table~\ref{tab:suspicion}). Overall, \textbf{\textsc{ScriptMind}}’s real-time, context-aware warnings promote stronger and more lasting cognitive defense than static alerts.

\section{Discussion}

\paragraph{Qualitative Analysis} Qualitative analyses revealed that fine-tuned models outperformed their zero-shot counterparts by accurately identifying detailed scam patterns that zero-shot models often misinterpreted or overlooked (Table~\ref{tab:conversation_examples}). They also achieved lower false positive and false negative rates, with more coherent scammer utterance predictions and rationale explanations, highlighting their strong potential for real-world deployment in localized scam detection systems (Appendix~\ref{Finetuning_Analysis}).

\paragraph{Design Direction for the Scam Detection} \textbf{\textsc{ScriptMind}} significantly heightened and sustained suspicion during scams. In particular, behavioral prediction of scammers effectively facilitated the transition from suspicion to conviction, demonstrating that LLMs can function as dynamic cognitive companions. Building on this insight, our proposed on-device LLM system continuously monitors scam dialogues, predicts deceptive intent across conversational stages, and provides adaptive notices to sustain user vigilance. A corresponding UI/UX prototype embodies this interactive flow, guiding the development of cognitively adaptive Scam Detection systems in the future(Appendix~\ref{UI}).

\paragraph{Ethical Considerations} LLM-based scam detection and on-device deployment entail inherent privacy and misuse risks. Scam datasets and models could be exploited by malicious attackers, so we implemented multiple safety measures to mitigate such threats. First, all experiments were conducted on de-identified and anonymized data derived from verified voice phishing cases, and no original audio or personally identifiable information was used. Second, to mitigate the risk of adversarial misuse, model training and inference were performed exclusively within a secure, closed police network, and the model itself was not released; only textual outputs were analyzed. Third, the system was designed as a human-in-the-loop decision-support tool rather than an autonomous surveillance mechanism. Finally, the human-subject study received institutional review board approval, and all procedures complied with applicable privacy regulations and emerging AI governance frameworks for high-impact public-sector AI systems(Appendix~\ref{safetymeasures}).

\section{Conclusion}
We presented \textbf{\textsc{ScriptMind}}, an integrated framework for crime script inference and cognitive evaluation in LLM-based social engineering scam detection. Unlike prior systems, it models the cognitive dynamics of user–AI interaction, connecting automated detection with human-centered defense. Experiments showed that fine-tuning improved accuracy, reduced false positives, and produced more interpretable explanations than baselines. Cognitive simulations revealed that \textbf{\textsc{ScriptMind}} interventions strengthened users’ suspicion, turning scam detection into an awareness-driven defense.

Despite these results, limitations remain. Emotional states could not be fully measured due to ethical constraints, and the study focused only on phone scams, excluding multimodal attacks. Future work will optimize on-device performance and expand beyond the Korean dataset. Overall, \textbf{\textsc{ScriptMind}} advances cognitively adaptive LLMs for scam prevention, showing how crime script inference can enhance model reasoning and awareness against evolving social engineering threats.

\section*{Limitations}

We validated a real-time, script-aware LLM model, identifying suspicion as a reliable cognitive marker. Despite uniformly high initial suspicion from ethical disclosure, a three-step validation confirmed the marker’s validity, the null effect of single warnings, and significant LLM effects at Stages 4–5. However, high baseline suspicion constrained affective measures such as anxiety and trust. Future work should incorporate multimodal sensing to capture subtler emotional responses.

We further validated the model’s predictive reasoning through crime script analysis and statistical evaluation, focusing on prosecutor-impersonation scams. Yet the findings remain limited to phone based cases. As social engineering evolves with deepfakes, voice cloning, and multimodal impersonation, future research should develop cross modal script for broader threat understanding.

We focused on enhancing users’ cognitive resilience rather than optimizing inference speed or latency for deployment. Nonetheless, fine-tuning 1–2B-parameter models showed practical efficiency and clear trade-offs compared to larger ones, highlighting their potential for lightweight implementation.

The data used in this study was constructed based on Korean prosecutor impersonation phone scam cases. For broader applicability, future research should extend to cases from other countries.

\section*{Acknowledgment}
This work was supported by the Institute of Information \& Communications Technology Planning \& Evaluation (IITP) grant funded by the Korea government (MSIT) (No. 2022-0-00653, Development of a Voice Phishing Information Collection, Processing, and Big Data–Based Investigation Support System).

\bibliography{custom}

\appendix

\clearpage
\section{UI and UX Design Result}
\label{UI}

Based on our experiment results demonstrating that \textbf{\textsc{ScriptMind}} enhances users’ cognitive awareness, we designed a system capable of efficiently detecting real-time social engineering scams that occur during phone calls in real-world device environments. This design concretizes the conceptual framework of \textbf{\textsc{ScriptMind}} from a practical perspective, providing the foundational operational structure for future research and development of more advanced real-time scam detection systems.

\begin{figure}[h]
    \centering
    \includegraphics[width=1\columnwidth]{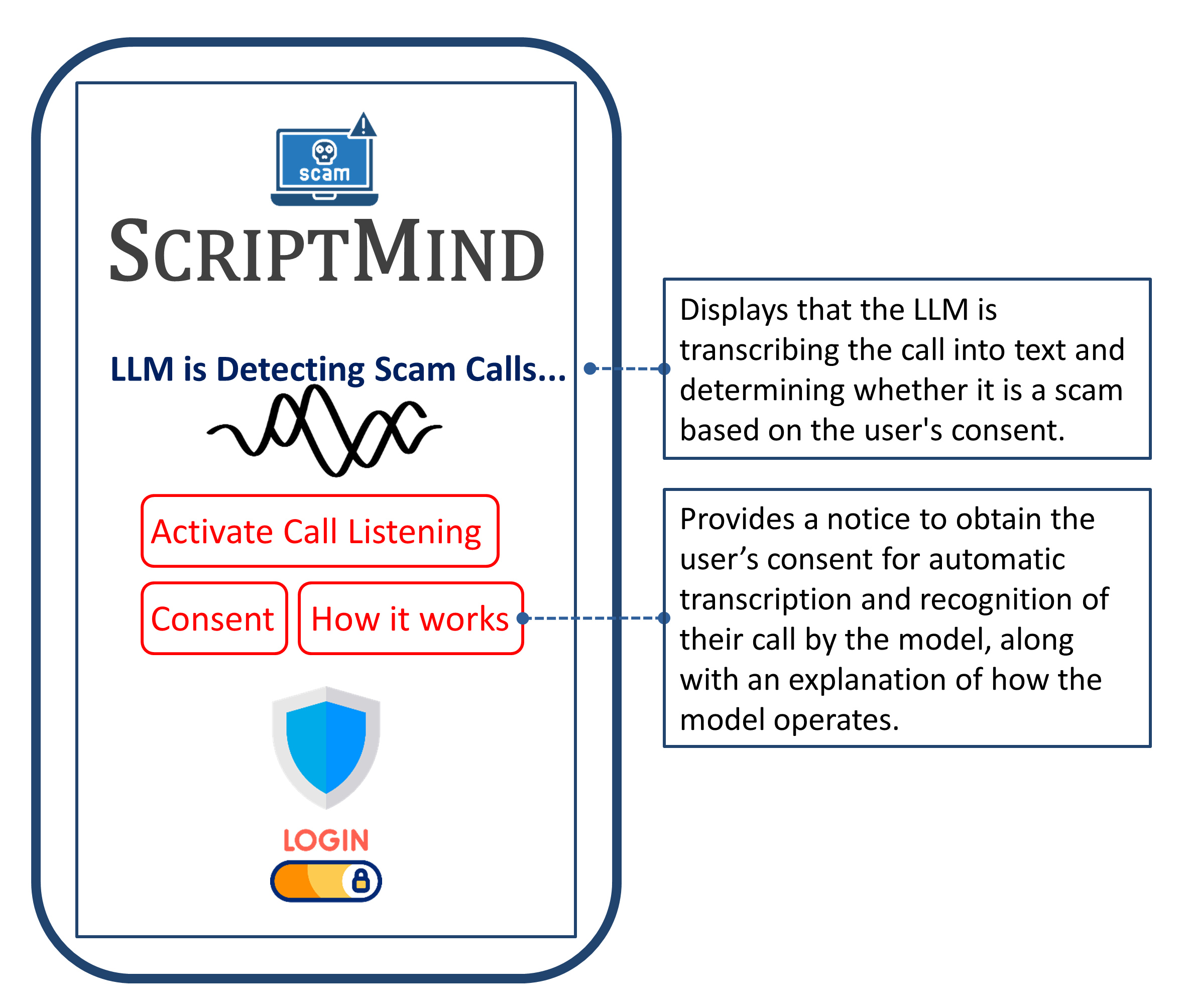} 
    \caption{User Consent Interface and Real-Time Transcription Workflow of \textbf{\textsc{ScriptMind}}}
    \label{fig:fig111}
\end{figure}

Figure~\ref{fig:fig111}\textasciitilde\ref{fig:fig333} illustrate the main operational flow of \textbf{\textsc{ScriptMind}}, assuming that it runs as a smartphone application. As shown in Figure~\ref{fig:fig111}, \textbf{\textsc{ScriptMind}} obtains the user’s explicit consent to automatically transcribe phone conversations (speech-to-text) and employs a LLM to analyze and display the likelihood of fraud for each conversational segment. When suspicious activity is detected, the system immediately displays a warning message, enabling the user to compare the model’s prediction with the actual dialogue and make an informed decision. Upon first launch or activation of the monitoring feature, a Consent modal appears to obtain user permission, and users can disable the function at any time via the settings menu, which instantly stops real-time analysis.

\begin{figure}[h]
    \centering
    \includegraphics[width=1\columnwidth]{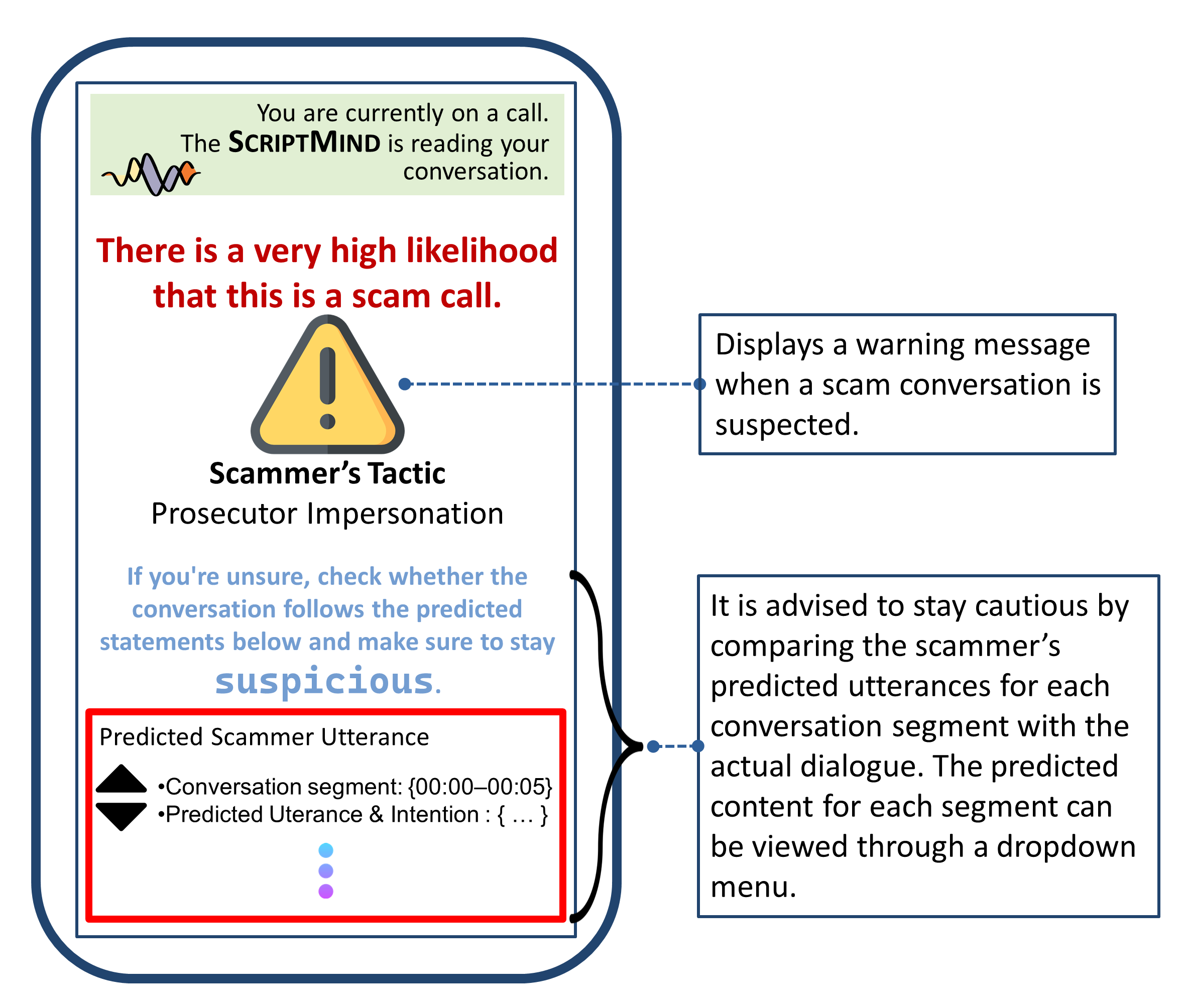} 
    \caption{LLM-Based Scam Recognition and Next-Utterance Prediction Display.}
    \label{fig:fig222}
\end{figure}

\begin{figure}[h]
    \centering
    \includegraphics[width=1\columnwidth]{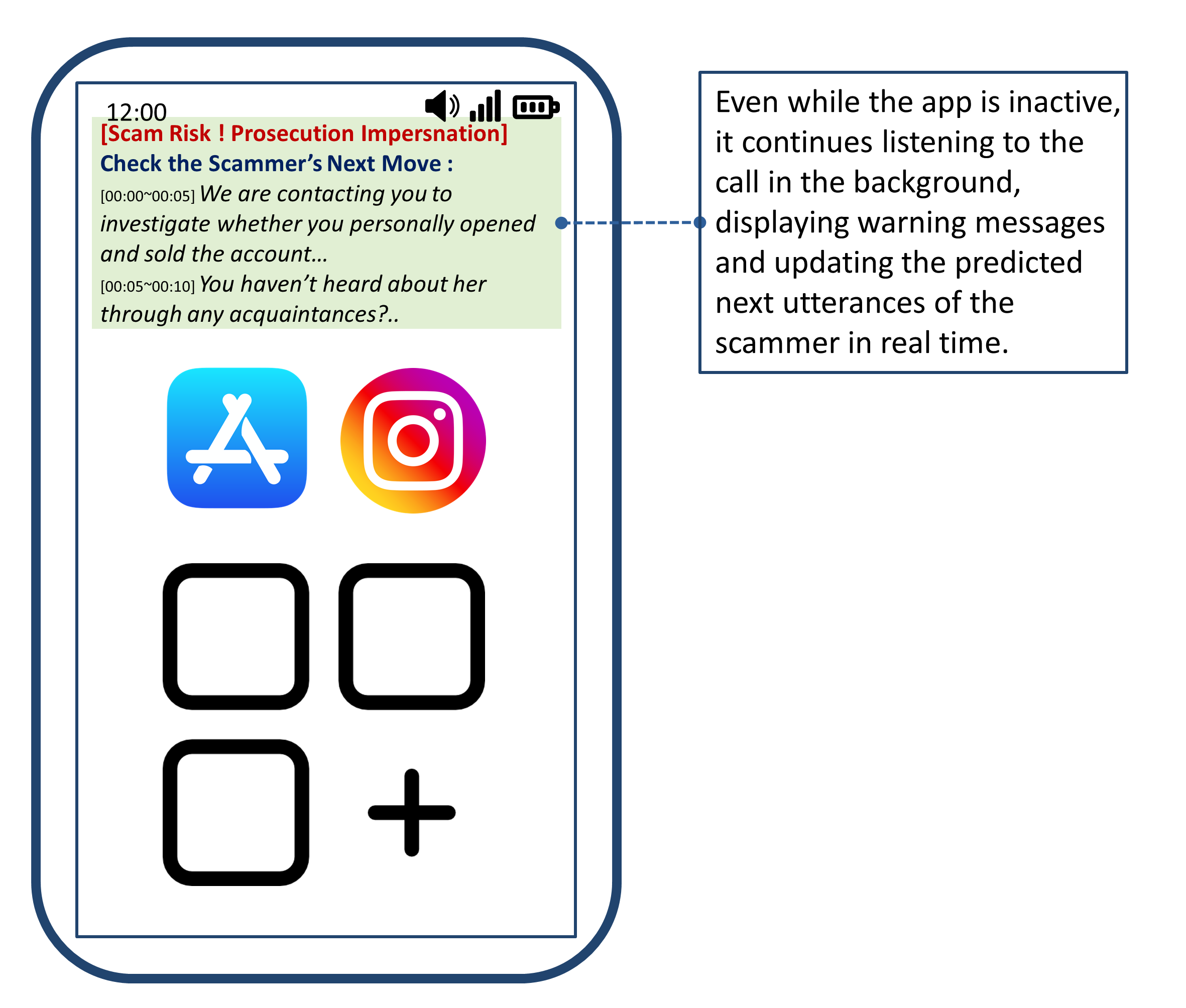} 
    \caption{Background Monitoring and Real-Time Scam Alert Notification}
    \label{fig:fig333}
\end{figure}

Next, as depicted in Figure~\ref{fig:fig222}, the LLM continuously analyzes the transcribed conversation in real time. When a scam is detected, \textbf{\textsc{ScriptMind}} displays the identified scam type along with one of its core outputs, the predicted next utterance of the scammer. Since model prediction may experience slight latency compared to the actual conversation, \textbf{\textsc{ScriptMind}} divides the dialogue into second-level segments so that users can scroll through and view the predictions for each segment. This design allows users, even after the call ends, to retrospectively recognize that “the model’s prediction was indeed correct,” thereby reinforcing their awareness and caution against scam tactics.

Finally, as shown in Figure~\ref{fig:fig333}, \textbf{\textsc{ScriptMind}} continues to function in the background even when the application is inactive. The system delivers alert notifications based on the model’s prediction results, allowing users to receive scam warnings in real time while communicating through speakerphone. It ensures continuous protection and real-time guidance without requiring active interaction.

\clearpage
\section{Scam Dataset}
\label{sec:appendix}

\subsection{Original Dataset}

\begin{tcolorbox}[
breakable,
width=\columnwidth,
colback=gray!2,
fonttitle=\bfseries,
coltitle=white,
boxrule=0.4pt,
arc=2mm,
left=2mm,
right=2mm,
top=1mm,
bottom=1mm,
enhanced
]
\textit{We define the overall scam conversation dataset as the original dataset \( D_o \),  
which contains conversation between users and scammers:}

{\normalsize
\[
\begin{aligned}
D_o &= \{ \mathcal{C} \}, \quad \mathcal{C} = \{ C^{(1)}, C^{(2)}, \dots, C^{(n)} \} \\
C^{(i)} &= \{ U^{\text{usr}}_t, U^{\text{scm}}_t \mid t = 1, 2, \dots, T_i \}
\end{aligned}
\]
\textit{Each case \( C^{(i)} \) consists of sequential utterances exchanged between a user and a scammer, where \(U^{\text{usr}}_t\) and \(U^{\text{scm}}_t\) denote the user’s and scammer’s utterances at turn \(t\), respectively,  
and \(T_i\) is the total number of turns.
}}

\end{tcolorbox}

\begin{table}[h!]
\centering
\small
\begin{adjustbox}{max width=\columnwidth} 
\begin{tabularx}{\columnwidth}{X}
\toprule
\textbf{Original Scam Conversation Example} \\
\midrule

This is the Seoul Central District Prosecutors' Office. \\

My name is Investigator Yoo Seowon from the Advanced Crime Investigation Division. \\

Yes, this is the Seoul Central District Prosecutors' Office. \\

Yes, the reason I’m calling is regarding an identity theft case involving Mr./Ms. [Name]. \\

Before I explain the details of the case, may I ask if you know a person named [Name]? \\

No, it’s a male from [Address], aged [Age]. \\

Just a moment. Recently, we arrested a financial fraud ring called “[Name],” and during the operation, we seized numerous credit cards and illegal bank accounts.\\
\bottomrule
\end{tabularx}
\end{adjustbox}
\caption{Example of scam conversation.}
\end{table}

\subsection{Initial Dataset}

\begin{tcolorbox}[
breakable,
width=\columnwidth,
colback=gray!2,
fonttitle=\bfseries,
coltitle=white,
boxrule=0.4pt,
arc=2mm,
left=2mm,
right=2mm,
top=1mm,
bottom=1mm,
enhanced
]
\textit{We define the Fraudulent Intent Interpretation (FII) dataset as an intent classification dataset derived from each conversation \( C^{(i)} \) in the original set. It maps subsets of scam-related utterances to corresponding intent labels.}

{\normalsize
\[
\begin{aligned}
D_I &= \{ (U^{(i)}_S, Y^{(i)}_I) \mid C^{(i)} \in \mathcal{C} \} \\
U^{(i)}_S &\subseteq C^{(i)}, \quad
Y^{(i)}_I \subseteq \mathcal{Y}_{\text{intent}} = \{ y_1, \dots, y_m \}
\end{aligned}
\]

\textit{Here, \( U^{(i)}_S \) represents a subset of utterances within a conversation \( C^{(i)} \),  
and \( Y^{(i)}_I \) denotes the corresponding subset of intent labels drawn from the intent label set \(\mathcal{Y}_{\text{intent}}\).}
}

{\normalsize
\[
D_o \supset \mathcal{C} \supset C^{(i)} \supset U^{(i)}_S, 
\quad
\mathcal{Y}_{\text{intent}} \supset Y^{(i)}_I
\]
}

\end{tcolorbox}

\begin{table}[h!]
\centering
\small 
\begin{tabularx}{\columnwidth}{lX}
\toprule
\texttt{"conversation"} &
``Right, so you’re saying that on May 23, 2016, at the [Address] branch—  
No, I didn’t. (So you didn’t personally open that account, correct?) Yes.  
So, to confirm again, you’re saying that you didn’t open the account yourself, correct?  
Yes. Since the account was seized during the illegal operation, we have frozen it to verify whether it was personally opened or fraudulently used.’’ \\
\midrule
\texttt{"intention"} &
\texttt{["3. Case introduction – (7) Verify whether the bank account was personally opened by the citizen",  
"6. Investigation recording – (3) Notify that the account has been frozen"]} \\
\bottomrule
\end{tabularx}
\caption{Example of an instance containing conversation and corresponding intentions from Law\&Order Dataset.}
\end{table}

\subsection{Scammer's Behavior Sequences}
\label{SBS}

\begin{tcolorbox}[
breakable,
width=\columnwidth,
colback=gray!2,
fonttitle=\bfseries,
coltitle=white,
boxrule=0.4pt,
arc=2mm,
left=2mm,
right=2mm,
top=1mm,
bottom=1mm,
enhanced
]
\textit{We further define a subset of scammer utterances from \( U^{(i)}_S \) that are explicitly annotated with intent labels, forming the Scammer-Based Subset (SBS) dataset.}

{\normalsize
\[
U^{(i)}_{\text{scm,int}} = \{ U^{\text{scm}}_t \in U^{(i)}_S \mid Y^{(i)}_I(U^{\text{scm}}_t) \neq \varnothing \}
\]
\textit{This subset includes only the scammer utterances for which corresponding intent annotations exist.}
}

{\normalsize
\[
D^{(i)}_{\text{scm}} =
\Big(
  \{ U^{\text{scm}}_1, U^{\text{scm}}_2,\dots \},
  \{ Y_1, Y_2,\dots \}
\Big)
\]
\textit{To represent these utterances and their intent labels as paired data, each conversation \( C^{(i)} \) produces a local mapping between scammer utterances and corresponding intent categories.}
}

{\normalsize
\[
D_{\text{SBS}} =
\{ (U^{\text{scm}}_t, Y_t) \mid U^{\text{scm}}_t \in U^{(i)}_S, \; Y_t \in Y^{(i)}_I \}
\]
or equivalently,
\[
D_{\text{SBS}} = \{ (U^{\text{scm}}_t, Y_t) \}
\]
\textit{where each pair \( (U^{\text{scm}}_t, Y_t) \) corresponds to a scammer utterance and its associated intent.}
}

\end{tcolorbox}

\begin{table*}[h]
\centering
\begin{adjustbox}{max width=\textwidth}
\scriptsize
\begin{tabularx}{\textwidth}{p{1cm} p{1.2cm} X p{3cm} p{3cm}}
\toprule
\textbf{Case ID} & \textbf{Speaker} & \textbf{Utterance} & \textbf{Scenario Classification} & \textbf{Intent Classification} \\
\midrule
001 & Scammer & The account was issued at [Address], opened in June and used until November. & 4. Case Involvement & Informing the citizen that objective evidence confirms their connection to the case. \\
001 & Scammer & Were you not aware of this account? & 3. Case Introduction & Checking whether the citizen knows about the account involved in the crime. \\
001 & Scammer & The reason I contacted you is to confirm whether you personally opened and sold these two accounts to the “[Name]” group for payment, or whether, like other victims, your identity was stolen. Our preliminary investigation did not find any evidence suggesting you colluded with “[Name]”, so we are contacting you under the assumption that your name was misused. & 4. Case Involvement & Verifying whether the person actually sold the account or was a victim of identity theft. \\
001 & Scammer & What’s important now is determining whether you are a victim or an accomplice in this case. At the scene, we found two bank accounts under your name — from [Bank Name] — that were used in this crime, and there are victims who suffered financial loss through those accounts. & 4. Case Involvement & Explaining that multiple people are involved in the crime, including both perpetrators and victims of identity theft. \\
\midrule
002 & Scammer & This is the Seoul Central District Prosecutors’ Office. & 2. Self-introduction & Stating the fake identity being impersonated. \\
002 & Scammer & I am Investigator Yoo Seowon from the Advanced Crime Investigation Division. & 2. Self-introduction & Stating the fake identity being impersonated. \\
002 & Scammer & Yes, this is the Seoul Central District Prosecutors’ Office. & 2. Self-introduction & Stating the fake identity being impersonated. \\
\bottomrule
\end{tabularx}
\end{adjustbox}
\caption{Examples of annotated scammer's behavior sequences with scenario and intent classification. All utterance categories are listed in Appendix~\ref{Scenario_Details}.}
\end{table*}

\subsection{Crime Script-Aware Inference Dataset}
\label{CSID}

\begin{tcolorbox}[
breakable,
width=\columnwidth,
colback=gray!2,
fonttitle=\bfseries,
coltitle=white,
boxrule=0.4pt,
arc=2mm,
left=2mm,
right=2mm,
top=1mm,
bottom=1mm,
enhanced
]
\textit{The Crime Script Inference Dataset (CSID) is constructed as a collection of conversational instances designed to model both predictive and interpretive aspects of scam communication. Each instance contains a partial dialogue segment, its scam-related label, the predicted next scam utterance, and an explanatory intent description.}

\vspace{0.6em}

{\normalsize
\[
\mathcal{D}_{\mathrm{CSID}} = \{ (U_S^{(i)}, Y_a^{(i)}, U_{t+1}^{(i)}, Y_{\mathrm{int}}^{(i)}) \}_{i=1}^{N}
\]
\textit{where \(U_S^{(i)}\) denotes a partial conversation within a single case,  
\(Y_a^{(i)} \in \{0,1\}\) indicates whether the segment represents a scam (1) or non-scam (0),  
\(U_{t+1}^{(i)}\) is the predicted next utterance of the scammer following \(U_S^{(i)}\),  
and \(Y_{\mathrm{int}}^{(i)}\) is a natural language description explaining the underlying intent of the conversation segment.}
}

\vspace{0.8em}

{\normalsize
\[
f_{\mathrm{CSID}} : U_S \;\mapsto\; (Y_a, U_{t+1}, Y_{\mathrm{int}})
\]
\textit{Here, the model \( f_{\mathrm{CSID}} \) learns to infer the likelihood of a scam,  
predict the scammer’s next utterance, and generate an intent-level explanation from a given dialogue segment.}
}

\end{tcolorbox}

\begin{table}[H]
\scriptsize
\begin{tabular}{p{4cm}|p{1cm}|p{1.4cm}}
\hline
\textbf{Conversation} & \textbf{Label} & \textbf{Explanation} \\ 
\hline
Do you have no knowledge about this at all? Alright, understood for now. Have you ever visited the [address] branch by any chance? This is the Seoul Central District Prosecutors’ Office. Yes. You’ve never been there, correct? Yes. The account we discovered was opened around August 2015 at the [address] branch. (What exact date?) It shows that it was opened around August 2015 at the [address] branch. That’s why I asked you about this earlier. In the past three years, have you ever lost any items such as your wallet or ID card that could lead to personal information leakage? According to our comparison with the relevant financial institution, this account is definitely registered under your name. & 
\texttt{scam} & 
\textbf{Next utterance:} “We are contacting you to investigate whether you personally opened the account and received payment for transferring it, or if you are a victim of identity theft.” \textbf{Rationale:} The scammer is attempting to confirm whether the victim sold the account or was impersonated. \\ 
\hline
Hello, is this Ms. Hwang Ga-eun? Yes, that’s me. Who is this? Hello, this is Sergeant Lee Cheol-soo from the Gangnam Police Station’s Traffic Department. Do you have a moment to talk? Yes, you said Traffic Department? What is this about? A complaint has been filed regarding your violation of the Road Traffic Act and dangerous driving resulting in injury. You are required to undergo an investigation for a drunk driving case that occurred on February 28, 2015. Ah, that case... I see. I’m very sorry. When should I come to the police station? Please let me know when you are available. I’ll schedule the investigation. How about this Friday morning? Friday morning works. Please come to our police station at 10 a.m. and ask for Sergeant Lee Cheol-soo at the Civil Affairs Office. Okay, I understand. I’ll come at 10 a.m. Alright, see you then. Please make sure to bring your ID. Yes, I’ll bring it. Thank you. Not at all. See you on Friday. Goodbye. Goodbye. & 
\texttt{non\_scam} & 
\textbf{Rationale:} This is a legitimate call from a police officer. The officer provides identification, clear instructions, and no suspicious requests. \\ 
\hline
\end{tabular}
\caption{An example from our completed Crime Script Inference Dataset (\textbf{CSID})}
\label{benign}
\end{table}

\begin{table}[ht]
\centering
\small
\begin{tabularx}{\textwidth}{lX}
\toprule
\textbf{Role} & \textbf{Instruction / Content} \\
\midrule
\textbf{SYSTEM} & You are an expert in detecting Korean phone scam conversations. Your output must strictly be a single JSON object. (No extra text outside the defined format.) \\
\midrule
\textbf{USER} & \{conversation\} \newline
(Follow the rules below.) \newline
- If the conversation is phone scam, set \texttt{label:"scam"} and fill in \texttt{next\_utterance} and \texttt{rationale}. \newline
- If it is not scam, return only \texttt{\{"label":"non\_scam"\}}. \\
\midrule
\textbf{Example Output (for a scam case)} &
\texttt{\{ \newline
\ \ "label": "scam", \newline
\ \ "next\_utterance": "Predicted next utterance of the scammer (1--2 sentences)", \newline
\ \ "rationale": "Current criminal intent: ... . Expected next criminal intent: ... . Evidence: ..." \newline
\}} \\
\midrule
\textbf{Task} &
Analyze the given conversation and return the result as JSON. \newline
\textbf{OUTPUT MUST BE VALID JSON. NO EXTRA TEXT.} \\
\midrule
\textbf{Conversation Example} &
\texttt{"Are you saying you have no knowledge of this at all? \newline
Alright, I understand. \newline
Have you, by any chance, visited the [address] branch before? \newline
This is the Seoul Central District Prosecutor’s Office. \newline
So, have you ever been to that branch? \newline
Yes. The bank account we found was opened around August 2015 at the [address] branch. \newline
That’s why we’re asking you. \newline
In the past three years, have you ever lost your wallet or any ID card that might have led to personal information leakage? \newline
After cross-checking with the financial institution, it is confirmed that the account was indeed opened under your name."} \\
\midrule
\textbf{Ground-truth Output} &
\texttt{"output": \{ \newline
\ \ "label": "scam", \newline
\ \ "next\_utterance": "The scammer’s next likely statement would be: 'We are contacting you to determine whether you personally opened and transferred this account for financial gain or if your identity has been stolen.'", \newline
\ \ "rationale": "The scammer currently aims to confirm the victim’s personal information leakage and is expected to next assess whether the victim is an active participant or a victim of identity theft." \newline
\}} \\
\bottomrule
\end{tabularx}
\caption{Example of LLM instruction for Korean social engineering scam detection}
\end{table}

\clearpage
\section{Behavior Sequence Analysis of Scammer's Utterances}
\label{SR_analysis}

Social engineering scams can be decomposed into step-by-step procedures through Crime Script Analysis~\cite{hutchings2015crime, loggen2022unraveling, choi2017voice, lwin2023supporting}, or strategically mapped into structured tables based on adversarial tactic models such as MITRE ATT\&CK~\cite{shin2022focusing, abo2024application}. We apply these analytical techniques to the scammer’s utterance sequences to identify where each utterance is positioned within the structured crime script of the scammer. Through this analysis, we capture both the psychological flow and tactical components of social engineering scams, providing a foundational basis for dataset labeling in training LLMs.


\noindent \textbf{Core Assumptions.} A key assumption in behavior sequence analysis of scammer's utterance is that typical tactics, such as those involving impersonation of prosecutors, follow a consistent pattern characterized by scripted dialogue structures, scenario progression, and intent-driven language. This assumption is supported by previous studies in crime script analysis, which have identified the step-by-step nature of social engineering scams~\cite{choi2017voice, lwin2023supporting}.

\noindent \textbf{Statistical Analysis.} Based on the structure and labels of the dataset, we statistically extracted recurring patterns in scammer's utterances along with their associated intentions. To analyze the relationships between these intentions, we calculated the transition frequencies between utterances labeled with each intent and derived Standardized Residual (SR) values~\cite{everitt2010cambridge}. 

SR is calculated as:

\begin{equation}
\small
\text{SR} = \frac{\text{Residual}}{\text{Standard Deviation of Prediction Error}}
\end{equation}

Where:

\begin{itemize}
\small
  \item \textbf{Residual} $= \text{Observed Value} - \text{Predicted Value}$
  \item \textbf{Standard Deviation of Prediction Error} reflects the uncertainty (variance) in the prediction for that specific observation.
\end{itemize}

Residuals with large absolute values are typically considered potential outliers~\cite{everitt2010cambridge}. In our study, we interpret such high SRs as indicative of repeated social engineering scam attempts following the same script, where an scammer consistently produces utterances that are more frequently observed than predicted in the dataset. This interpretation aligns with analytical approaches used in other domains, such as murder pattern analysis~\cite{marono2020behaviour}, where recurring behavioral patterns beyond statistical expectation are treated as significant indicators of intentional or scripted criminal activity.

\begin{figure}[h]
\centering
\includegraphics[width=1\columnwidth]{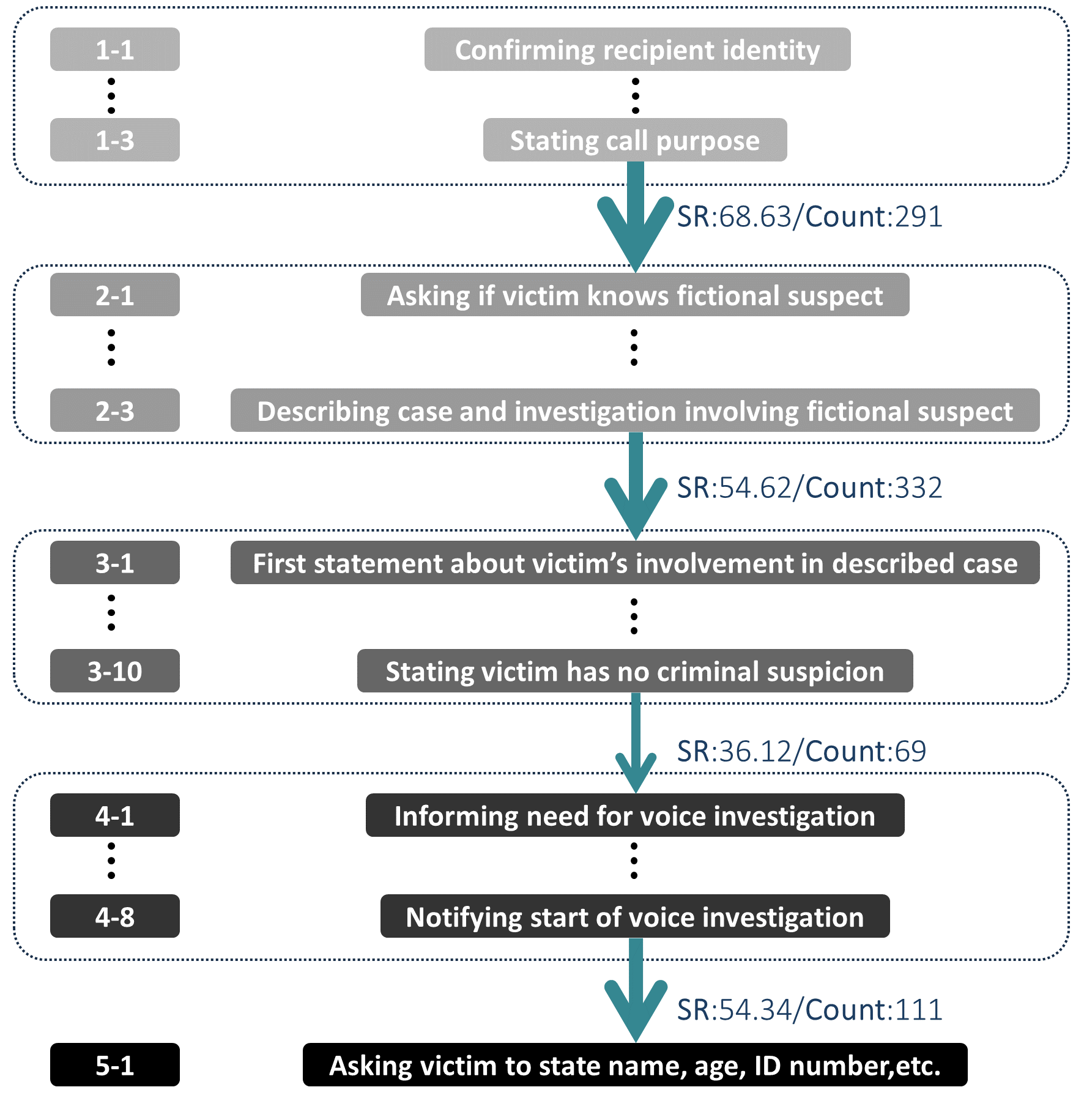}
\caption{A transition network of utterance sequences observed within 571 phone scam conversation data points. The arrows represent the preceding and immediately following utterance sequences, respectively. Each node number corresponds to the classified intent of the scammer’s utterance, as shown in the numbered utterance mapping in Appendix~\ref{Scenario_Details}. The SRs and frequencies of each sequence are listed in Table~\ref{tab:script_transitions}.}\label{network}
\end{figure}

\noindent \textbf{Analysis Results.} The analysis revealed notable similarities in utterance patterns even across different scenarios, suggesting that language models trained on the structured stages of social engineering scam, rather than on the broader contextual content, may be more effective in detecting and explaining attacker behavior.

\noindent \textbf{Labeling Validity.} The dataset used in this study consists of 571 phone scam cases, each annotated with intent labels by two professional crime profilers. The inter-rater agreement, measured using Cohen’s Kappa coefficient, reached 0.91, indicating a very high level of consistency and validating the reliability of the intent annotation process.

\noindent \textbf{Structural Patterns of Scam Scripts.} In the next phase, we analyzed the sequential order of utterances and intents across different cases to examine the structural regularities of scam scripts. To do so, we calculated the transition frequencies between utterances labeled with specific intents and derived the Standardised Residuals (SR) scores. Figure~\ref{network} presents a network visualization of the top 28 utterance sequences with the highest SR values. The results revealed that utterance sequences with an SR score of 2 or higher appeared frequently and consistently across scenarios, indicating a high degree of structural organization in scam dialogues. As shown in Table~\ref{tab:script_transitions}, there are 28 transition sequences with an SR value of 20 or higher, with the highest SR value reaching 74.19. Additionally, a total of 251 sequences were identified as statistically significant transitions with SR values of 2 or above. The validity of this structure is supported by the observation that identical strategic utterance sequences appeared repeatedly across different cases with a frequency well beyond random chance. In other words, social engineering scams tend to follow a standardized script in which the same intents and strategies are executed in a fixed order.

\begin{table}[h!]
\centering
\renewcommand{\arraystretch}{0.85}
\begin{tabular}{ccccc}
\hline
\textbf{No.} & \textbf{From ID} & \textbf{To ID} & \textbf{Count} & \textbf{SR} \\
\hline
1  & 5-(1) & 5-(2) & 135 & 74.19 \\
2  & 1-(3) & 2-1   & 291 & 68.63 \\
3  & 2-(1) & 2-(2) & 11  & 68.01 \\
4  & 1-(2) & 1-(3) & 273 & 67.61 \\
5  & 2-(1) & 2-(2) & 254 & 61.01 \\
6  & 5-(2) & 5-(3) & 106 & 57.91 \\
7  & 1-(1) & 1-(2) & 243 & 57.03 \\
8  & 5-(6) & 5-(8) & 91  & 55.78 \\
9  & 2-(3) & 3-1   & 332 & 54.62 \\
10 & 4-(8) & 5-1   & 111 & 54.34 \\
11 & 5-(5) & 5-(6) & 94  & 52.21 \\
12 & 4-(6) & 4-(7) & 127 & 41.20 \\
13 & 3-(1) & 2-(6) & 166 & 37.43 \\
14 & 5-(8) & 5-(9) & 36  & 37.26 \\
15 & 2-(2) & 2-(3) & 189 & 37.15 \\
16 & 3-(10)& 4-1   & 69  & 36.12 \\
17 & 5-(3) & 5-(5) & 69  & 34.33 \\
18 & 4-(7) & 4-(8) & 74  & 32.81 \\
19 & 3-(2) & 2-(7) & 134 & 31.95 \\
20 & 4-(1) & 3-(11)& 58  & 29.46 \\
21 & 4-(2) & 4-(6) & 68  & 25.30 \\
22 & 5-(4) & 5-(5) & 39  & 24.48 \\
23 & 2-(4) & 3-(10)& 68  & 24.09 \\
24 & 5-(5) & 5-(4) & 36  & 22.44 \\
25 & 5-(9) & 5-(8) & 23  & 22.31 \\
26 & 2-(6) & 2-(7) & 98  & 21.39 \\
27 & 5-(4) & 5-(6) & 30  & 20.47 \\
28 & 2-(3) & 2-(4) & 122 & 20.46 \\
\hline
\end{tabular}
\caption{Transition matrix of labeled utterances. Each node ID corresponds to the classified intent of the scammer’s utterance, as shown in the numbered utterance mapping in Appendix~\ref{Scenario_Details}.}
\label{tab:script_transitions}
\end{table}

\section{Model Selection}
\label{modelselction}

To verify the effectiveness of our \textbf{\textsc{ScriptMind}}-based social engineering scam detection system performing \textbf{CSIT}, we selected a diverse set of models as shown in Table~\ref{tab:model_summary}. First, since real-world scam detection must operate in restricted environments such as users’ on-device systems to ensure privacy protection, we included lightweight sLLMs (1\textasciitilde2B parameters). Next, we selected 7\textasciitilde11B parameter models to evaluate the feasibility of deploying them in secure intranet servers of public-sector organizations (e.g., police) using limited GPU resources. Finally, large-scale commercial models (e.g., GPT-4) with over 11B parameters were incorporated as baselines, allowing us to compare the latest high-performance reasoning capabilities. Since our \textbf{CSID} dataset is Korean-language based, we primarily focused on Korean-tuned models, while also evaluating multilingual models to examine their cross-lingual adaptability.

\begin{table}[h]
\centering
\scriptsize
\resizebox{\columnwidth}{!}{%
\begin{tabularx}{\columnwidth}{
>{\raggedright\arraybackslash}m{0.6cm}  
>{\raggedright\arraybackslash}m{1.3cm}  
>{\raggedright\arraybackslash}m{2.8cm}  
>{\centering\arraybackslash}m{1.5cm}    
}
\toprule
\textbf{Scale} & \textbf{Category} & \textbf{Model} & \textbf{Deployment} \\
\midrule

\multirow{3}{*}{\textbf{1--2B}} 
& Multilingual sLLM & Llama-3.2-1B-Instruct & On-device phone \\
\cline{2-3}
& Korean sLLM & Exaone-2B &  \\
&  & MIDM-mini &  \\

\midrule

\multirow{6}{*}{\textbf{7--11B}} 
& Multilingual sLLM & Llama-3.1-8B-Instruct & Closed intranet server \\
&  & SOLAR-10.7B-Instruct &  \\
\cline{2-3} 
& Korean sLLM & EEVE-Korean-Instruct-10.8B &  \\
&  & Exaone-7B &  \\
&  & MIDM-base &  \\

\midrule

\multirow{2}{*}{\textbf{>11B}} 
& Commercial LLM & chatgpt-4o-latest & No \\
&  & gemini-2.0-flash & \\
&  & Clade-10B-Instruct & \\

\bottomrule
\end{tabularx}%
}
\caption{Evaluation models categorized by parameter scale and deployment environment.}
\label{tab:model_summary}
\end{table}

\clearpage
\section{Cognitive Evaluation(CSED) Settings}
\label{cognitive_experiment}

\subsection{Research Question Formulation}
\label{aim_scope}
The experimental design of our study is grounded in a key research question: \textbf{\textit{Can real-time LLM-based scam detection serve as truly effective intervention tools?}} Unlike a single-point phishing, chat based online scam is a continuous and interactive process in which the victim’s psychological state evolves over time~\cite{martin2021social, kumarage2025personalized}. In particular, levels of suspicion are not static but tend to fluctuate depending on the phase of the scam~\cite{han2024voice}. These dynamic cognitive shifts raise important questions about the adequacy of traditional black-box detection models, which typically offer one-time warnings with limited contextual feedback. In response, there is increasing interest in LLMs that can deliver iterative and interpretable alerts throughout the interaction, aligning more closely with the user's changing cognitive state. Against this backdrop, our study designed and conducted a controlled cognitive experiment using a prosecutor impersonation voice phishing scenario to evaluate the effectiveness of real-time, LLM-based interventions.

Our phone scam scenario was selected as a representative case of a sophisticated scam, often executed through highly coordinated scripts by organized call center operations~\cite{choi2017voice}. It was chosen for two primary reasons. First, phone scam unfolds gradually through a sequence of conversations, making it particularly difficult to determine the optimal timing for detection-based intervention technologies~\cite{yu2024korean, choi2017voice}. Second, because detection relies solely on the spoken content of the conversation, distinguishing between legitimate and fraudulent calls is inherently challenging, thereby increasing the risk of false positives~\cite{shen2025warned}. Once technical filters are bypassed, the final decision, such as whether to hang up the call, rests entirely with the user. These characteristics make this scenario especially suitable for evaluating the effectiveness of real-time LLM-based scam detection and for gaining insights into the central research questions.

We formulate the following research questions:

\begin{itemize}
\item \textbf{RQ1:} How do recipients' emotional and cognitive responses change over the course of a phone scam conversation?
\item \textbf{RQ2:} Are current AI detection technologies effective in helping users recognize scam?
\item \textbf{RQ3:} Does detection and explanation of scams using LLMs enhance user awareness more effectively than conventional single-warning detection models?
\end{itemize}

\begin{figure*}[h!]
\centering
\includegraphics[width=0.95\textwidth]{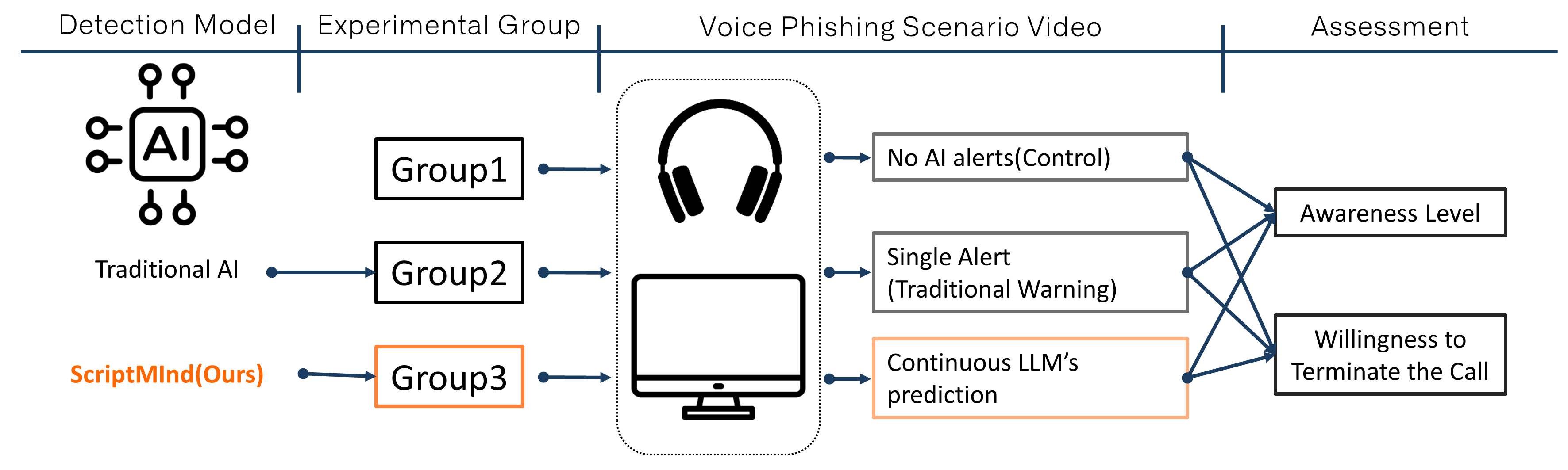}
\caption{Description of the Psychological Experimental Stimuli Simulating a Phone Scam Scenario}\label{psychology}
\label{overviewfig}
\end{figure*}

\begin{table*}[h!]
\centering
\renewcommand{\arraystretch}{0.9}
\small
\begin{tabular}{@{}p{6cm} p{9cm}@{}}
\toprule
\textbf{Stage} & \textbf{Content Description and Example Dialogue} \\
\midrule
Stage 1 (Introduction) & 
The attacker introduces themselves and presents a fabricated case. 
\newline 
\textit{“This is Investigator Kim Young-jae from the Seoul Central District Prosecutors’ Office.”} \\

Stage 2 (Allegation) & 
The recipient is falsely implicated in a crime. 
\newline 
\textit{“We recently arrested a financial fraud ring including someone named [Name], and during the seizure, we found large amounts of cash, cloned credit cards, and bankbooks—including two accounts under your name from [Bank 1] and [Bank 2].”} \\

Stage 3 (Recorded Investigation) & 
The attacker explains that the call is being recorded for investigative purposes. 
\newline 
\textit{“Since your accounts were found at the crime scene, if you believe you're a victim, you’ll need to formally prove it. We’re currently conducting voice-recorded phone interviews for suspected victims.”} \\

Stage 4 (Financial Information Request) & 
The attacker demands sensitive banking details. 
\newline 
\textit{“To freeze any unauthorized accounts and prevent further harm, please state which banks you currently use legitimately. For depositor protection registration, could you also confirm the current balance of your [Bank Name] account as of today?”} \\

Stage 5 (Legal Consequences Notification) & 
The recipient is threatened with potential legal repercussions. 
\newline 
\textit{“We’ll send a subpoena to your home address. Please review it and visit our office once you receive it.”} \\
\bottomrule
\end{tabular}
\caption{Phone Scam Stimulus Stages with Example Dialogues}
\label{tab:stimuli}
\end{table*}

\subsection{Experiment Design}
\label{design}
\noindent \blackcircnum{1} \textbf{Conditions.} To evaluate the effectiveness and user acceptance of an AI-based scam system, a three-group experimental study was designed using a simulated voice phishing call scenario grounded in a structured scam script. Prior to the main simulation experiment, we first analyzed a corpus of voice phishing data to develop the detection-oriented LLM described in the following sections. Details of the scenario analysis are provided in Appendix~\ref{Scenario_Details}. The scenario followed a five-stage crime script structure, with participants instructed to listen and respond to the stimuli in real time. These five stages were derived from a crime script analysis of actual voice phishing cases, and are summarized in Table~\ref{tab:stimuli}. Depending on the assigned experimental condition, participants received varying types and intensities of AI generated alerts. This design aimed to examine how the presentation of AI predictions influences participants’ perceptions and behavioral responses, such as their intention to terminate the call.

\begin{table}[h!]
\centering
\renewcommand{\arraystretch}{1.2}
\small
\begin{tabular}{@{}p{1.1cm} p{2.2cm} p{3.5cm}@{}}
\toprule
\textbf{Group} & \textbf{Condition} & \textbf{AI Intervention} \\ \midrule
Group 1 & Control & No AI based alerts provided \\
Group 2 & Single Warning & A single alert at the most suspicious stage (financial info request) \\
Group 3 & \textbf{\textsc{ScriptMind}} LLM warning & Continuous prediction of attacker next utterance \\ \bottomrule
\end{tabular}
\caption{Experimental Conditions}
\label{tab:conditions}
\end{table}

As shown in Table~\ref{tab:conditions}, Group 1 (control condition) received no AI-based alerts during the experiment. In contrast, two experimental groups were formed where AI model interventions were introduced. Group 2 was assigned the ``single warning alert'' condition, modeled after approaches developed in previous studies. Group 3 was assigned a newly developed condition in which an \textbf{Script-Aware LLM} continuously presented the ``predicted next SE threat'' as the attacker speech progressed.

\noindent \blackcircnum{2} \textbf{Stimulus Methods.} Figure~\ref{overviewfig} provides an overview of the entire experimental procedure and the method of stimulus presentation. As shown in Table~\ref{tab:stimuli}, all participants were exposed to the same five-stage voice phishing video stimulus. Each of the five stages consisted of 4 to 12 segmented utterances, with a total of 40 utterances presented as stimuli across all stages. The scenario was based on a commonly used psychological manipulation tactic in voice phishing, in which the perpetrator impersonates a prosecutor and falsely accuses the target of involvement in a criminal case. 

Before the stimulus began, participants were given the following instruction:\\
\textit{"The content you are about to see may be either a phone scam attempt or a legitimate notice from a public prosecutor’s office.”}\\ This prompt served as a tool to elicit participants’ judgments about the authenticity of each utterance in a realistic setting, allowing us to quantitatively measure their level of suspicion at each stage.

The experiment was conducted in the same conference room, where participants independently watched the video and completed the survey using tablet PCs and stereo headphones. All groups heard the same audio stimulus, while only the visual and auditory warning cues varied across conditions to ensure internal validity.

During the stimulus, the scammer's utterances were presented through both voice and on-screen text. After each stage, participants were asked(Appendix~\ref{questions}): \begin{enumerate} \item Whether they believed the utterance was part of a scam call. \item How suspicious or anxious they felt after being exposed to the content. \end{enumerate}

To simulate realistic decision-making under time pressure, participants were instructed to respond promptly, with limited time allocated for each response. Depending on the assigned condition, the presentation of visual and auditory alerts varied: the control group received no warnings; the Single Warning group (Group 2) received a visual alert during Stage 4 (Financial Information Request); and the \textbf{\textsc{ScriptMind}} LLM warning group (Group 3) was shown an AI-generated sentence predicting the scammer next utterance as a visual cue, displaying a warning message—“Warning!! This is a scam call”—along with the logo of the Korean National Police Agency, accompanied by an auditory alert tone. The presentation of real-time prediction outputs was constructed by selecting sentences deemed accurate from the predictions generated in real time by the developed LLM model, based on pre-constructed scam scripts.

\begin{table}[h!]
\centering
\renewcommand{\arraystretch}{0.8} 
\small
\begin{tabular}{@{}p{1.6cm} p{0.6cm} p{0.6cm} p{0.6cm} p{0.6cm} p{1.4cm}@{}}
\toprule
\textbf{Group} & \textbf{20--29} & \textbf{30--39} & \textbf{40--49} & \textbf{50--59} & \textbf{Total} \\ \midrule
Group 1  & 8 & 7 & 8 & 7 & 30 \\
Group 2  & 7 & 8 & 7 & 8 & 30 \\
Group 3  & 8 & 7 & 8 & 7 & 30 \\
Total \\ (Preliminary) & 23 (+2) & 22 (+2) & 23 (+2) & 22 (+2) & 90 (+8) \\
\bottomrule
\end{tabular}
\caption{Number of Participants by Group and Age}
\label{group}
\end{table}

\noindent \blackcircnum{3} \textbf{Participant Recruitment.} As shown in Table~\ref{group}, a total of 98 adults aged 20 to 59 were recruited, evenly distributed across four age groups by decade and assigned to three experimental conditions through stratified random sampling. While G*Power analysis suggested a minimum of 34 participants per group for sufficient power, this study ensured robustness through a bootstrap based ANOVA and repeated measures design.

To ensure the reliability of the study and prevent data contamination, participants recruited based on pre-defined criteria were automatically assigned to condition groups. After inputting age information, each participant was randomly assigned to a condition within the experimental platform.

\begin{itemize}
    \item[\textbf{1)}] \textbf{Inclusion Criteria}: Individuals who voluntarily consented to participate after receiving an explanation of the study's purpose.
    \item[\textbf{2)}] \textbf{Exclusion Criteria}: Individuals who had participated in a survey or experiment within the past six months; those employed at financial institutions or law enforcement/judicial agencies; and those working in fields related to research such as marketing, market research, journalism, or broadcasting.
\end{itemize}

\subsection{Evaluation Questions}
\label{questions}
\noindent At each utterance stage, participants were presented with the following identical set of questions.

\noindent \textbf{Q1.} \textit{Who do you think this speaker is: an authority (e.g., investigator) or a scammer?}
\begin{itemize}
    \item 1–3: Investigator 
    \item 4: Not Sure 
    \item 5–7: Scammer
\end{itemize}

\noindent \textbf{Q2.} \textit{Emotional Evaluation – Please check the item that best describes your current feeling:}
\begin{itemize}
  \item \textbf{Importance:}
  \begin{itemize}
    \item 1–3: Not important at all 
    \item 4: Neutral 
    \item 5–7: Very important
  \end{itemize}
  \item \textbf{Relevance:}
  \begin{itemize}
    \item 1–3: Not relevant to me 
    \item 4: Neutral 
    \item 5–7: Highly relevant
  \end{itemize}
  \item \textbf{Anxiety:}
  \begin{itemize}
    \item 1–3: Not anxious at all 
    \item 4: Neutral 
    \item 5–7: Very anxious
  \end{itemize}
\end{itemize}

\subsection{Statistical Analysis}
\label{statistics_method}
We conducted quantitative statistical analyses to examine differences in perception, emotional response, and behavioral intention during scam call scenarios, based on AI warning types and call stages. Statistical analyses were performed using the \textit{JAMOVI} software, employing repeated measures ANOVA, one-way ANOVA, and independent samples \textit{t}-tests as the primary analytical methods. The significance level was set at $\alpha = .05$.
\begin{itemize}
    \item To examine how recipients' psychological responses as addressed in \textbf{RQ1}, including suspicion of fraud, anxiety, and perceived personal relevance, change over the course of a scam call, the five call stages were treated as repeated measures factors. A repeated measures ANOVA was conducted to analyze differences in psychological variables across stages, and Greenhouse-Geisser corrections were applied in cases where the assumption of sphericity was violated.

    \item To investigate the impact of the conventional AI detection method, namely a single warning message, on recipients’ scam recognition and their intention to terminate the call as addressed in \textbf{RQ2}, an independent samples \textit{t}-test and one-way ANOVA were conducted to compare differences between the single warning condition (Group 2) and the control condition without any warning (Group 1).

    \item To evaluate the effectiveness of the LLM-based real-time utterance prediction model compared to traditional methods as addressed in \textbf{RQ3}, three experimental conditions (\textbf{\textsc{ScriptMind}} LLM warning, single warning, and control group) were set as independent variables. A one-way ANOVA was conducted to assess their effects on scam recognition, intention to terminate the call, and attitudes toward AI intervention. Additionally, a mixed-design repeated measures ANOVA was performed to examine the interaction between AI alert condition and stage.
\end{itemize}

\subsection{Scam Details in Cognitive Experiment}
\label{Scenario_Details}
\subsection*{1. Identity Confirmation \& Introduction}
\begin{itemize}
  \item \textbf{(1)} Confirming recipient identity
  \begin{itemize}
      \item Ex) \textit{Hello, is this [Name]?}
  \end{itemize}
  \item \textbf{(2)} Stating impersonated identity
    \begin{itemize}
      \item Ex) \textit{Hello, this is Investigator Kim Young-jae from the Seoul Central District Prosecutors’ Office. Is now a good time to talk?}
  \end{itemize}
  \item \textbf{(3)} Stating call purpose
      \begin{itemize}
      \item Ex) \textit{I'm contacting you regarding a few confirmations about your personal data breach.}
     \end{itemize}

\end{itemize}

\subsection*{2. Case Introduction}
\begin{itemize}
  \item \textbf{(1)} Asking if victim knows fictional suspect
        \begin{itemize}
      \item Ex) \textit{Do you happen to know someone named Kim Sang-sik from Ilsan, Gyeonggi Province?}
     \end{itemize}
  \item \textbf{(2)} Asking about suspect's address, age, etc.
        \begin{itemize}
      \item Ex) \textit{He’s a former civil servant, a 47-year-old man. Have you ever heard about him through acquaintances?}
     \end{itemize}
  \item \textbf{(3)} Describing case and investigation involving fictional suspect
        \begin{itemize}
      \item Ex) \textit{We have arrested a financial fraud syndicate led by Kim Sang-sik.}
     \end{itemize}
  \item \textbf{(4)} Forming suspicion about account used in crime
        \begin{itemize}
      \item Ex) \textit{During the seizure, multiple bankbooks and IDs under your name (from Kookmin Bank and Shinhan Bank) were found. Are you aware of these accounts?}
     \end{itemize}
  \item \textbf{(5)} Disclosing bankbook purchase through testimony
        \begin{itemize}
      \item Ex) \textit{According to the statement made by [Name], when they purchased the bank account, they primarily used internet banking, transferring money into the account in their own name before completing the purchase.}
     \end{itemize}
  \item \textbf{(6)} Asking if victim knows the account
        \begin{itemize}
      \item Ex) \textit{This account was used in a crime that resulted in a victim. Are you aware of these accounts?}
     \end{itemize}
  \item \textbf{(7)} Confirming if victim opened the ghost account
        \begin{itemize}
      \item Ex) \textit{Did you, then, open [Bank Name] and [Bank Name] accounts under your name around January 27, 2016, through [Address]?}
     \end{itemize}
\end{itemize}

\subsection*{3. Case Involvement}
\begin{itemize}
  \item \textbf{(1)} Statement about victim's involvement in the case
          \begin{itemize}
      \item Ex) \textit{At the scene of the arrest, a large amount of cash, cloned credit cards, and bank accounts under borrowed names were seized. Among these items, bank accounts from [Bank Name] and [Bank Name] registered under your name were identified.}
     \end{itemize}
  \item \textbf{(2)} Objectively stating victim's link to case
            \begin{itemize}
      \item Ex) \textit{When we checked the issuance date of those accounts, it showed July 14, 2022, from Yeongdeungpo branch.}
     \end{itemize}
  \item \textbf{(3)} Confirming identity theft
  \begin{itemize}
      \item Ex) \textit{Have you ever received any message or contact about your personal data being leaked to financial firms or shopping malls?}
     \end{itemize}
  \item \textbf{(4)} Confirming whether it was theft or actual sale
    \begin{itemize}
      \item Ex) \textit{We contacted you to verify the misuse of bankbooks opened under your name.}
     \end{itemize}
  \item \textbf{(5)} Asking if victim sold or transferred account
      \begin{itemize}
      \item Ex) \textit{Have you ever transferred your bank account to another person?}
     \end{itemize}
  \item \textbf{(6)} Stating need for proof of victimization
        \begin{itemize}
      \item Ex) \textit{Just because your name is on the account doesn’t mean we see you as the criminal. We do consider you a possible victim, but we need proof.}
     \end{itemize}
  \item \textbf{(7)} Warning that sale/transfer leads to punishment
          \begin{itemize}
      \item Ex) \textit{If you did transfer your bank account, you may be subject to punishment under Article 10, Section 90 of the Act on the Punishment of Transfer of Personal Financial Information.}
     \end{itemize}
  \item \textbf{(8)} Pressuring that account was created by victim
            \begin{itemize}
      \item Ex) \textit{When we checked the issuance date of those accounts, it showed July 14, 2022, from Yeongdeungpo branch.}
     \end{itemize}
  \item \textbf{(9)} Explaining many involved, including victims
              \begin{itemize}
      \item Ex) \textit{This case currently involves approximately 180 individuals nationwide. Among them are people who either opened bank accounts and sold them or were victims whose identities were stolen.}
     \end{itemize}
  \item \textbf{(10)} Stating victim has no criminal suspicion
              \begin{itemize}
      \item Ex) \textit{Based on our investigation, you have no criminal history and verified identity, so we are contacting you in advance.}
     \end{itemize}
  \item \textbf{(11)} Informing of investigation via voice testimony
                \begin{itemize}
      \item Ex) \textit{We're here to assist your statement as a victim.}
     \end{itemize}
  \item \textbf{(12)} Notifying that proof of victimization is required
  \begin{itemize}
      \item Ex) \textit{Since both of your bank accounts were found at the crime scene, if you believe you are a victim, it is essential that you provide proof to establish your status as a victim.}
     \end{itemize}
  \item \textbf{(13)} If proven victim, informing of compensation
    \begin{itemize}
      \item Ex) \textit{If you are able to prove that you are a victim and it is confirmed that these individuals withdrew money from your account, the state can provide compensation for the loss.}
     \end{itemize}
  \item \textbf{(14)} Notifying that prosecution is investigating
      \begin{itemize}
      \item Ex) \textit{We’re not contacting you from a local police station or an insurance company, right? You understand where we’re calling from, correct? This is an official investigation by a government agency—the Seoul Central District Prosecutors’ Office.}
     \end{itemize}
  \item \textbf{(15)} Warning of severe penalty for false statements
        \begin{itemize}
      \item Ex) \textit{The entire investigation process is being recorded, so if you are aware of any details regarding this case but provide false statements or attempt to conceal information, you may be subject to more severe legal penalties.}
     \end{itemize}
\end{itemize}

\subsection*{4. Preparation for Voice Investigation}
\begin{itemize}
  \item \textbf{(1)} Informing need for voice investigation
        \begin{itemize}
      \item Ex) \textit{Since there’s no direct suspicion against you, we’ll proceed with a simplified voice-recorded investigation.}
     \end{itemize}
  \item \textbf{(2)} Inducing agreement to participate
          \begin{itemize}
      \item Ex) \textit{For now, we will only record the parts that you are aware of as evidence. Do you agree to the recording?}
     \end{itemize}
  \item \textbf{(3)} Prohibiting victim from revealing investigation
            \begin{itemize}
      \item Ex) \textit{And since you, [Name], are currently in the position of an interviewee under investigation, you do not have the right to disclose or discuss any details related to this case until your status as a victim has been verified. Understood?}
     \end{itemize}
  \item \textbf{(4)} Telling victim their accounts will be tracked
            \begin{itemize}
      \item Ex) \textit{We are currently conducting a joint investigation with [Agency Name], and we will be performing account tracing under your name. If the accounts with [Bank Name] and [Bank Name] were opened without your knowledge, there is a possibility that other undiscovered accounts may exist as well.}
     \end{itemize}
  \item \textbf{(5)} Telling victim to note impersonated info
            \begin{itemize}
      \item Ex) \textit{First of all, are you able to take notes? Since I’m the investigator in charge of your case, let me go over my affiliation and name again. Please get ready to write it down.}
     \end{itemize}
  \item \textbf{(6)} Notifying voice record will be submitted to court
            \begin{itemize}
      \item Ex) \textit{This will be submitted to court, so if there are background noises or third-party voices, it won't be accepted as evidence. Please take caution.}
     \end{itemize}
  \item \textbf{(7)} Setting environment for call
              \begin{itemize}
      \item Ex) \textit{Are you currently at home or at work? I asked because third-party voices can invalidate the recording.}
     \end{itemize}
  \item \textbf{(8)} Notifying start of voice investigation
  \begin{itemize}
      \item Ex) \textit{Alright, we’ll start the recording.}
     \end{itemize}
\end{itemize}

\subsection*{5. Voice Investigation}
\begin{itemize}
  \item \textbf{(1)} Asking victim to state name, ID, etc.
    \begin{itemize}
      \item Ex) \textit{Hello, I’m Investigator Kim Young-jae. Please state your name and age for the record.}
     \end{itemize}
  \item \textbf{(2)} Re-checking knowledge of people/accounts/crime links
  \begin{itemize}
      \item Ex) \textit{Do you know Kim Sung-sik, a 47-year-old man residing in Ilsan, Gyeonggi Province?}
     \end{itemize}
  \item \textbf{(3)} Notifying account freezing
  \begin{itemize}
      \item Ex) \textit{These two accounts were frozen to prevent further damage. Do you know what freezing means?}
     \end{itemize}
  \item \textbf{(4)} Notifying further freezing if other banks found
  \begin{itemize}
      \item Ex) \textit{Are you aware that any newly found accounts may result in penalties and freezing?}
     \end{itemize}
  \item \textbf{(5)} Confirming normal banks used
  \begin{itemize}
      \item Ex) \textit{If you use other bank accounts beyond those you’ve declared, please state only the name of the legitimate bank to prevent them from being frozen as illegal.}
     \end{itemize}
  \item \textbf{(6)} Confirming number/purpose of bank accounts
  \begin{itemize}
      \item Ex) \textit{At [Bank Name], how many accounts and for what purposes would you normally have under your name? You don’t have any accounts related to savings plans, housing subscriptions, funds, stocks, or cryptocurrency, correct?}
     \end{itemize}
  \item \textbf{(7)} Checking cash held in victim’s account
  \begin{itemize}
      \item Ex) \textit{By “freezing,” we mean you can’t use the account at all. Do you understand?}
     \end{itemize}
  \item \textbf{(8)} Confirming balance held in account
  \begin{itemize}
      \item Ex) \textit{For depositor protection registration, we need to verify today’s balance at [Bank Name].}
     \end{itemize}
  \item \textbf{(9)} Warning about punishment if amount differs
  \begin{itemize}
      \item Ex) \textit{If the reported balance differs by over 1 million KRW, the account will be considered suspicious, frozen, and may lead to an arrest warrant.}
     \end{itemize}
  \item \textbf{(10)} Informing of next steps after call
  \begin{itemize}
      \item Ex) \textit{Thank you for your cooperation. The first hearing will be held next Wednesday.}
     \end{itemize}
  \item \textbf{(11)} Threatening in-person summon if victim refuses to participate
  \begin{itemize}
      \item Ex) \textit{First, we will send a summons to your residence. Once you receive it, please appear in person at our office as instructed.}
     \end{itemize}
\end{itemize}

\clearpage
\section{Ethical Considerations}

\subsection{Dataset and Model}
\label{safetymeasures}

The data used in this study was derived from the \textsc{Fraudulent Scenario Completion} task, one of the benchmark tasks included in the officially released Law\&Order dataset, made available via github by a policy researcher affiliated with the Korean National Police Agency~\cite{lawandorder2025}. The original source data for this task consists of 571 voice phishing cases recorded between 2015 and 2019, all of which included verified voice recordings. Each case was transcribed using speech-to-text processing, and personally identifiable information, such as names, bank account numbers, and addresses, was either removed or anonymized to ensure that the dataset contained no personal data. For the purposes of this study, the original audio files were not used. Instead, the experiments were conducted using synthetic audio recordings, generated by professional actors reading the transcriptions.

The sLLM model used in the experiment was developed to predict the intent behind the scammer’s utterances, utilizing a generative language modeling approach. Given the potential risk of adversarial misuse, such as the model being exploited to generate scam content, both model training and inference were conducted exclusively within a closed, internal network at the Korean National Police Agency, which originally provided the dataset. The model itself was not released as open source; only the model outputs, in the form of text, were used in the experimental setting.

\subsection{Human Experiment}
\label{Ethics}
This research was reviewed and approved by the Central Institutional Review Board of Korea (IRB No. P01-202408-01-045), and written informed consent was obtained from all participants. All responses were collected anonymously, no personal data was stored or retained, and the entire experimental process adhered to ethical standards for research involving human subjects. Participant recruitment was conducted between August 26 and September 6, 2024. Eligible participants were pre-screened based on predefined inclusion criteria and were contacted via SMS with a URL containing an information sheet and consent form detailing the study purpose, procedures, potential risks and benefits, data protection measures, and voluntary participation policy. Participants who consented were randomly assigned to one of three conditions and were provided with a tablet and headphones to complete the task. Each participant viewed a simulated voice phishing scenario video and responded to stage-specific survey items over the course of approximately 30 to 40 minutes. All response data were collected via a secure electronic survey system in real time.

\subsection{Privacy Concerns of On-Device LLMs for Scam Conversation Analysis}

\subsubsection{Misuse as a Surveillance Tool}

Although scam conversation datasets can help build models that detect and prevent fraud, there is a potential concern that such data might be misused for automated surveillance. However, this concern is mitigated by multiple safeguards. Law enforcement agencies, telecom companies, and social media platforms are all bound by existing privacy and communication protection laws that strictly prohibit unauthorized monitoring of citizens' private communications. Moreover, these organizations already possess their own user data and cannot legally repurpose it for surveillance without consent or judicial oversight.

\subsubsection{Operation in Secure, Closed Environments}

Our dataset and models are developed and operated within the closed police network, ensuring that no data is used for targeting individuals. The system functions solely as an internal decision-support tool for investigators, with all judgments ultimately made by human officers. All preprocessing and experiments occur within this restricted environment and are not accessible to external parties.

\subsubsection{Legal and Regulatory Safeguards}

Countries such as South Korea, the UK, and members of the EU have established or proposed AI governance frameworks, for example, Korea’s Artificial Intelligence Basic Act, that classify police-developed AI systems as “high-impact AI.” Such systems are legally required to undergo committee review, data management oversight, user protection planning, and risk mitigation. These international regulatory trends collectively ensure that on-device LLMs analyzing scam conversations cannot be used for broad or invasive surveillance.

\clearpage
\section{Finetuning Results Analysis}
\label{Finetuning_Analysis}

\subsection{Correlation between LLM-as-a-Judge and Human Evaluation}

\begin{table}[htbp]
\centering
\small
\begin{tabular}{l c}
\hline
\textbf{Pair} & \textbf{Correlation} \\
\hline
LLM-finetuning vs Human1-finetuning & 0.850$^{***}$ \\
LLM-zeroshot vs Human1-zeroshot & 0.830$^{***}$ \\
Human1-zeroshot vs Human2-zeroshot & 0.826$^{***}$ \\
Human1-finetuning vs Human2-finetuning & 0.819$^{***}$ \\
LLM-finetuning vs Human2-finetuning & 0.810$^{***}$ \\
LLM-zeroshot vs Human2-zeroshot & 0.782$^{***}$ \\
\hline
\end{tabular}
\caption{Results of Correlation Analysis on LLM-as-a-Judge and Human Evaluation. Pearson correlation coefficients were calculated. Statistical significance was set at $p < 0.05$. $^{***}p < 0.001$.}
\label{tab:correlation}
\end{table}

\begin{table*}[ht]
\centering
\begin{tabularx}{\textwidth}{X}
\toprule
\textbf{Instruction / Content} \\
\midrule
You are an expert evaluator for phone scam scenario predictions. Your task is to compare the model's predicted next utterance with the correct ground truth utterance. Rate the prediction \textbf{STRICTLY} based on whether it conveys the \emph{same phishing situation} or meaning as the ground truth, not merely based on text similarity. \newline

Give a score between \texttt{0.00} and \texttt{1.00} (two decimal places): \newline
- \texttt{1.00} means the prediction fully matches the meaning and intent of the ground truth (same phishing situation described). \newline
- \texttt{0.00} means the prediction is completely different or unrelated. \newline
- Intermediate values (e.g., \texttt{0.45}, \texttt{0.72}) represent partial semantic overlap or situational similarity. \newline

Output \textbf{only the numeric score}, no explanation. \\
\midrule
\end{tabularx}
\caption{LLM instruction example for automated evaluation of scam scenario predictions}
\label{judgeinstruction}
\end{table*}

We evaluated the quality of LLM responses for \textbf{CSIT} using the LLM-as-a-Judge framework. To validate its reliability, we analyzed the correlation between two human experts’ ratings and the automatic scores on 200 randomly sampled instances from the test set. The analysis employed the Pearson correlation coefficient, with statistical significance set at p < 0.05. We conducted independent analyses for 200 zeroshot and 200 finetuning evaluation instances, following the same procedure. Both human experts and the LLM were instructed to assess responses solely based on the given golden answers, with the LLM providing decimal scores between 0 and 1 and humans using a 7-point Likert scale. The evaluation prompt is presented in Table~\ref{judgeinstruction}.

As shown in Table~\ref{tab:correlation}, the results indicate a strong correlation between human and LLM-based evaluations, consistently observed across both zeroshot and finetuning settings. The correlations were statistically significant, suggesting that automated evaluation by LLMs can serve as a valid alternative for assessing other models.

\subsection{Effect of sLLM Fine-tuning}
We conducted a detailed comparison between the zero-shot and fine-tuned performances of each sLLM to assess the impact of \textbf{\textsc{ScriptMind}} finetuning. As shown in Table~\ref{tab:zs_results}, all seven models demonstrated improvements across scam detection accuracy, next-utterance quality, and rationale generation. On average, the Accuracy and F1 score of scam detection increased by 0.28 and 0.19, respectively, while the False Positive Rate decreased by approximately 0.28, indicating a notable reduction in misclassification. Although the False Negative Rate varied by model, with some showing slight increases(e.g. Llama 3.1 8B and SOLAR 10.7B), this trend may reflect a more conservative classification tendency when models jointly learned benign data, causing them to label borderline scam instances as non-scam. In contrast, both Next Utterance and Rationale scores improved substantially (0.24, 0.16), suggesting that \textbf{\textsc{ScriptMind}} finetuning enhanced contextual understanding and explanatory quality in scam-related dialogue modeling.

\begin{table*}[h!]
\centering
\scalebox{0.9}{
\begin{tabular}{|l|ccc|c|c|}
\hline
\multicolumn{1}{|c|}{\multirow{2}{*}{\textbf{Model}}} &
\multicolumn{3}{c|}{\textbf{Scam Detection}} &
\multicolumn{1}{c|}{\textbf{Next Utterance}} &
\multicolumn{1}{c|}{\textbf{Rationale}} \\ \cline{2-6}
\multicolumn{1}{|c|}{} &
\textbf{ACC} & \textbf{F1} & \textbf{FP / FN} &
\textbf{LLM-as-a-Judge} & \textbf{LLM-as-a-Judge} \\ \hline

\texttt{Llama-3.2-1B-Instruct} & 0.36 & 0.36 & -0.18 / -0.19 & 0.35 & 0.40 \\ \hline
\texttt{EXAONE-3.5-2.4B-Instruct} & 0.36 & 0.29 & -0.25 / -0.10 & 0.23 & 0.22 \\ \hline
\texttt{Midm-2.0-Mini-Instruct} & 0.26 & 0.23 & -0.20 / -0.06 & 0.22 & 0.24 \\ \hline
\texttt{Llama-3.1-8B-Instruct} & 0.30 & 0.09 & -0.49 / 0.19 & 0.22 & 0.00 \\ \hline
\texttt{SOLAR-10.7B-Instruct} & 0.21 & 0.10 & -0.33 / 0.12 & 0.28 & 0.06 \\ \hline
\texttt{EEVE-Korean-Instruct-10.8B} & 0.27 & 0.24 & -0.20 / -0.08 & 0.26 & 0.28 \\ \hline
\texttt{Midm-2.0-Base-Instruct} & 0.18 & 0.04 & -0.28 / 0.09 & 0.11 & -0.07 \\ \hline

\end{tabular}
}
\caption{Performance improvement of each model before and after fine-tuning. Each value represents the performance gain of the fine-tuned model compared to its zero-shot baseline.}
\label{tab:zs_results}
\end{table*}

In addition, we qualitatively analyzed the improvement in scam detection performance achieved through \textbf{\textsc{ScriptMind}} fine-tuning by examining actual prediction cases.

\paragraph{(1) Enhanced Understanding of Scam Scenarios}  
As shown in the first row of Table~\ref{tab:conversation_examples}, the EEVE model fine-tuned with \textbf{\textsc{ScriptMind}} demonstrates a clear understanding of a typical scam scenario in which the scammer falsely claims that \textit{“the user’s bank account is linked to a criminal case.”} Consequently, in other similar cases where the scammer states that \textit{“it is necessary to verify whether the user opened the account themselves or is a victim of identity theft,”} the fine-tuned model accurately predicts such repetitive and characteristic scam utterances and provides detailed explanations of their deceptive intent. In contrast, the zero-shot model merely repeats the scammer’s words or offers only a superficial description such as \textit{“the scammer is trying to steal personal information.”}  

\paragraph{(2) Reduction of False Negatives}  
As illustrated in the second row of Table~\ref{tab:conversation_examples}, the \textbf{\textsc{ScriptMind}} fine-tuned EEVE model successfully identifies deceptive intent even in conversations that appear ordinary at first glance. For instance, an utterance like \textit{“Do you know Mr. XX?”} may sound like a casual question, but in a \textit{prosecutor impersonation scam scenario}, it serves as a classic tactic to gain trust by referring to a fictional criminal figure. The fine-tuned model accurately recognized this contextual cue and classified the dialogue as a scam, whereas the zero-shot model misclassified it as a normal conversation. This finding highlights the importance of enabling early-stage scam detection to prevent further interaction and potential victimization, emphasizing the necessity of learning subtle contextual cues underlying scam communication.  

\paragraph{(3) Reduction of False Positives}  
While minimizing false negatives is important, reducing false positives is an even more critical challenge in scam detection. If legitimate communications from law enforcement, such as requests for appearance issued by the police or prosecution, are mistakenly classified as scams, the reliability of the detection system would be severely compromised, potentially disrupting legitimate criminal justice operations. As presented in the third row of Table~\ref{tab:conversation_examples}, the \textbf{\textsc{ScriptMind}} fine-tuned EEVE model correctly classifies such lawful law-enforcement scenarios as non-scam, whereas the zero-shot model erroneously labels them as scam. This demonstrates the necessity of incorporating both scam and benign data in a balanced fine-tuning process to enable the model to distinguish subtle linguistic and contextual differences between fraudulent and legitimate communications. Through this approach, the model can precisely capture the characteristics of deceptive speech while maintaining high accuracy in recognizing genuine public communication.

\begin{table*}[h!]
\centering
\scriptsize
\begin{tabular}{|p{2.5cm}|p{0.7cm}|p{1.6cm}|p{1.6cm}|p{1.6cm}|p{1.6cm}|p{1.6cm}|p{1.6cm}|}
\hline
\textbf{Input Conversation} & \textbf{Golden Label} & \textbf{Golden Answer \tiny(Next Utterance)} & \textbf{Golden Answer \tiny(Rationale)} & \textbf{Finetuned Ans \tiny(Next Utterance)} & \textbf{Finetuned Ans \tiny(Rationale)} & \textbf{Zeroshot Ans \tiny(Next Utterance)} & \textbf{Zeroshot Ans \tiny(Rationale)} \\ 
\hline

“Do you have no knowledge about this at all?  
Alright, understood.  
Have you ever visited the [address] branch?  
This is the Seoul Central District Prosecutors’ Office.  
Have you ever been there?  
The account discovered was opened around August 2015 at the [address] branch.  
That’s why I asked you earlier.  
In the past three years, have you ever lost any items like your wallet or ID that could lead to personal data leakage?  
According to our comparison with the relevant financial institution, the account is definitely registered under your name.” 
& \colorbox{black!30}{\texttt{scam}} 
& “The scammer’s next expected utterance: We are contacting you to investigate whether \colorbox{gray!20}{you personally} \colorbox{gray!20}{opened and sold} \colorbox{gray!20}{the account} for payment or if you were a \colorbox{gray!20}{victim of identity} \colorbox{gray!20}{theft.}” 
& “The scammer currently aims to \colorbox{cyan!20}{verify whether} \colorbox{cyan!20}{the victim’s} \colorbox{cyan!20}{identity was} \colorbox{cyan!20}{stolen} and will next ask whether the victim \colorbox{cyan!20}{sold the account} \colorbox{cyan!20}{or was} \colorbox{cyan!20}{impersonated.}” 
& “The scammer’s next expected utterance: The reason we are contacting you today is to determine whether \colorbox{gray!20}{you personally} \colorbox{gray!20}{opened the} \colorbox{gray!20}{accounts} and sold them for profit, or \colorbox{gray!20}{whether your} \colorbox{gray!20}{identity was} \colorbox{gray!20}{stolen.}” 
& “The scammer aims to inform the citizen about \colorbox{cyan!20}{the case} \colorbox{cyan!20}{objectively} and will next ask whether they \colorbox{cyan!20}{sold or were} \colorbox{cyan!20}{impersonated.}” 
& “The scammer’s next expected utterance: Have you ever visited the [address] branch?” 
& “Current scammer intent: to obtain the user’s financial information.  
Next expected intent: to collect ID or banking information, as the scammer is inquiring about the user’s address and transactions.” \\ 
\hline

“There was an issue, so we’re contacting you.  
We have a few questions about this case—are you available for a call now?  
Do you know a woman named [Name], born in [Address], aged [Age]?” 
& \colorbox{black!30}{\texttt{scam}} 
& “The scammer’s next expected utterance: So, \colorbox{gray!20}{you haven’t} \colorbox{gray!20}{heard about her} through any acquaintances or groups?  
We are asking because the prosecution \colorbox{gray!20}{recently arrested} \colorbox{gray!20}{a financial fraud} \colorbox{gray!20}{ring led by} \colorbox{gray!20}{[Name]}, during which we seized multiple bank accounts” 
& “The scammer currently seeks to check if the victim \colorbox{cyan!20}{recognizes a} \colorbox{cyan!20}{fake suspect} name and will next explain \colorbox{cyan!20}{the fabricated} \colorbox{cyan!20}{case} and the citizen’s alleged involvement.” 
& “The scammer’s next expected utterance: \colorbox{gray!20}{You don’t know} \colorbox{gray!20}{this person?}  
We recently \colorbox{gray!20}{arrested a} \colorbox{gray!20}{financial crime} \colorbox{gray!20}{ring led by} \colorbox{gray!20}{[Name].}” 
& “The scammer aims to confirm whether the user knows \colorbox{cyan!20}{the fake suspect} and will next \colorbox{cyan!20}{describe the} \colorbox{cyan!20}{crime} and investigation details.” 
& \multicolumn{2}{p{3.2cm}|}{\centering \colorbox{red!15}{\textit{Misclassified as non-scam.}}} \\
\hline

“Is this Mr. Yoon Young-sik?  
Yes, who is this?  
Hello, this is Sergeant Lee Cheol-soo from the Gimhae Police Department’s Traffic Investigation Team.  
Do you have a moment to talk?  
Yes, what is this about?  
We received a report of drunk driving on August 25, 2018.  
We need to investigate this matter.  
When can you come to the station?  
This Friday at 3 p.m. would work.  
Please bring your ID when you visit.” 
& \multicolumn{3}{p{3.9cm}|}{\centering \colorbox{black!30}{\texttt{non-scam}}}
& \multicolumn{2}{p{3.2cm}|}{\centering \colorbox{red!15}{\textit{“Correctly classified as non-scam.”}}}
& \multicolumn{2}{p{3.2cm}|}{\centering \colorbox{red!15}{\textit{Misclassified as scam.}} \colorbox{red!5}{Incorrectly predicted next utterance} : “The officer informs the citizen of a drunk-driving investigation and reminds them to bring their ID.”, \colorbox{red!5}{Incorrectly predicted intent}: to deceive Mr. Yoon under the guise of a police officer to obtain personal ID information and to use the ID and personal data for identity theft.} \\ 
\hline

\end{tabular}
\caption{Examples of input conversations, ground-truth (golden) answers, and model predictions (finetuned vs. zero-shot). The examples illustrate model behavior in next utterance generation and rationale explanation for both scam and non-scam dialogues.}
\label{tab:conversation_examples}
\end{table*}

\clearpage
\section{Cognitive Evaluation Results Analysis}
\label{researchquestions}

\subsection{Summary of Experiment Result Analysis}
Because real scam call experiments are ethically infeasible, participants in our simulation were informed that the conversation might be fraudulent, which naturally elevated their initial suspicion levels. To mitigate this limitation and ensure the validity of the findings, we applied a rigorous multi-step analytical framework to capture the genuine cognitive effects of LLM interventions.

First, we statistically examined whether participants’ suspicion varied meaningfully across scam stages. Alongside suspicion, we also tracked anxiety and perceived relevance to validate suspicion as a sensitive cognitive indicator. Results revealed that while anxiety and relevance showed no significant stage-wise differences, suspicion decreased across Stages 1–3 and sharply increased at Stage 4. A repeated measures ANOVA confirmed that these differences were statistically significant (Appendix~\ref{RQ1}), verifying that suspicion functions as a more dynamic and diagnostic psychological marker for scam detection.

Next, to assess the impact of a traditional single warning, we compared mean perceived suspicion scores across the entire script, as single warnings are not tied to specific conversational stages. Although the single-warning group reported slightly higher suspicion than the control group, the difference was not statistically significant (Appendix~\ref{RQ2}). This suggests that a one-time alert may momentarily raise awareness but cannot sustain cognitive resistance throughout a strategically structured social engineering scam.

Finally, we analyzed the stage-wise effects of the LLM intervention. A significant interaction between stage and group (Table~\ref{tab:anova_results}) indicated that suspicion patterns were not driven merely by call progression but by the type of warning received. Stage-wise ANOVAs further showed that the LLM Warning group exhibited the highest suspicion at Stages 4 and 5, with statistically significant between-group differences (Table~\ref{tab:stage_group_stats}). These findings indicate that LLM interventions effectively sustain and amplify suspicion, especially during critical stages involving pressure or financial solicitation, whereas the control and single-warning groups showed slower or incomplete recovery in awareness.

\subsection{RQ1: suspicion evolves, but emotion persists throughout scam stages} 
\label{RQ1}

\begin{figure}[t!]
\centering%
\includegraphics[width=0.48\textwidth]{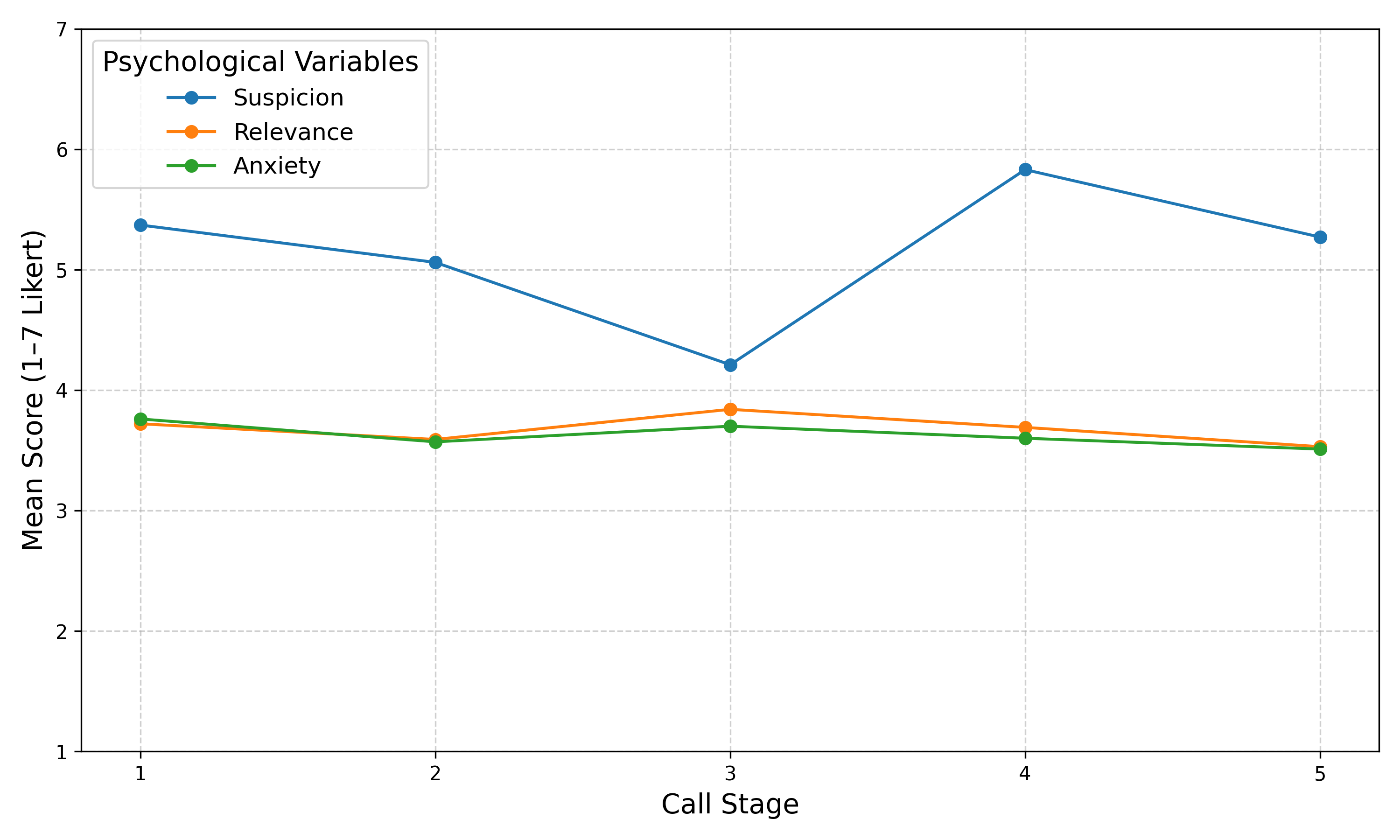}
\caption{Mean Trends of Psychological Reactions}\label{graph1}
\end{figure}

To address \textbf{RQ1}, we examined how recipients' psychological reactions, specifically suspicion, perceived relevance, and anxiety change. The results show that suspicion levels temporarily decreased and reached their lowest point at Stage 3, followed by a sharp increase at Stage 4. In contrast, perceived relevance and anxiety remained relatively stable across stages. The means and standard deviations for each psychological variable across the five stages are presented in Table~\ref{tab:stage_means}, and these patterns are also visualized in Figure~\ref{graph1}.

\begin{table}[h!]
\renewcommand{\arraystretch}{1.2}
\centering
\footnotesize
\begin{tabular}{p{1cm} p{2.1cm} p{1.4cm} p{1.4cm}}
\hline
\textbf{Stage} & \textbf{Suspicion (M ± SD)} & \textbf{Relevance} & \textbf{Anxiety} \\
\hline
Stage 1 & 5.37 ± 1.89 & 3.72 ± 1.87 & 3.76 ± 1.93 \\
Stage 2 & 5.06 ± 1.90 & 3.59 ± 1.89 & 3.57 ± 1.91 \\
Stage 3 & 4.21 ± 2.06 & 3.84 ± 1.97 & 3.70 ± 1.89 \\
Stage 4 & 5.83 ± 1.64 & 3.69 ± 2.03 & 3.60 ± 2.06 \\
Stage 5 & 5.27 ± 2.08 & 3.53 ± 2.00 & 3.51 ± 2.05 \\
\hline
\end{tabular}
\caption{Descriptive Statistics of Reactions}
\label{tab:stage_means}
\end{table}

The results of the repeated measures ANOVA support this finding. As shown in Table~\ref{tab:anova_by_variable}, there was a statistically significant effect of stage on suspicion($F$(4, 356) = 23.20, $p < .001$). In contrast, no significant differences across stages were observed for perceived relevance($F$(4, 356) = 1.45, $p = .217$), or anxiety($F$(4, 356) = 1.10, $p = .359$).

\begin{table}[h!]
\renewcommand{\arraystretch}{1}
\centering
\small
\begin{tabular}{p{1.5cm} p{1.5cm} p{1.5cm} p{1.7cm}}
\hline
\textbf{Variable} & \textbf{F(4, 356)} & \textbf{P} & \textbf{Partial $\eta^2$} \\
\hline
Suspicion & 23.20 & $< .001$ & .207 \\
Relevance & 1.45  & .217     & .016 \\
Anxiety   & 1.10  & .359     & .012 \\
\hline
\end{tabular}
\caption{Repeated Measures ANOVA with Effect Sizes. Partial $\eta^2$ represents the effect size. Sphericity assumption was violated for all variables, but the Greenhouse Geisser corrected results yielded consistent patterns.}
\label{tab:anova_by_variable}
\end{table}

\begin{table}[h!]
\renewcommand{\arraystretch}{1.4}
\centering
\small
\begin{tabular}{p{1cm} p{1cm} p{1.3cm} p{1cm} p{1.3cm}}
\hline
\textbf{Stage 1} & \textbf{Stage 2} & \textbf{Mean Difference} & \textbf{P} & \textbf{Significant} \\
\hline
Stage 1 & Stage 2 & –0.31 & .814  & No \\
Stage 1 & Stage 3 & –1.16 & $< .001$ & Yes \\
Stage 1 & Stage 4 &  0.47 & .479  & No \\
Stage 1 & Stage 5 & –0.10 & .997  & No \\
Stage 2 & Stage 3 & –0.84 & .028  & Yes \\
\hline
\end{tabular}
\caption{Pairwise comparisons of suspicion. We used Tukey’s HSD Test. Mean differences reflect the direction and magnitude of change between stages.}
\label{tab:stage_comparisons}
\end{table}

Post-hoc comparisons using Tukey’s HSD test were also conducted for the suspicion variable. As shown in Table~\ref{tab:stage_comparisons}, significant differences were observed between Stage 1 and Stage 3 ($p$ $<$ .001), and between Stage 2 and Stage 3 ($p$ = .028), indicating that suspicion was significantly lower at Stage 3 than in the earlier stages. 

These findings suggest that participants showed initial suspicion during Stages 1–2, which briefly declined in Stage 3, likely due to persuasive scammer cues, then sharply increased from Stage 4 as pressure and financial demands escalated. In contrast, relevance and anxiety remained relatively stable, indicating that suspicion may be a more sensitive and dynamic marker for detecting scam.

\subsection{RQ2: single interventions failed to produce significant
cognitive effects in complex social engineering scam} 
\label{RQ2}

To address \textbf{RQ2}, which examines the effect of a traditional AI-based detection method on recipients’ scam awareness and call termination intention, independent samples t-tests and one-way ANOVA were conducted to compare the single warning and no-warning groups. Awareness was measured as the average perceived suspicion score across all stages of the script.

\begin{figure}[h!]
\centering
\includegraphics[width=0.5\textwidth]{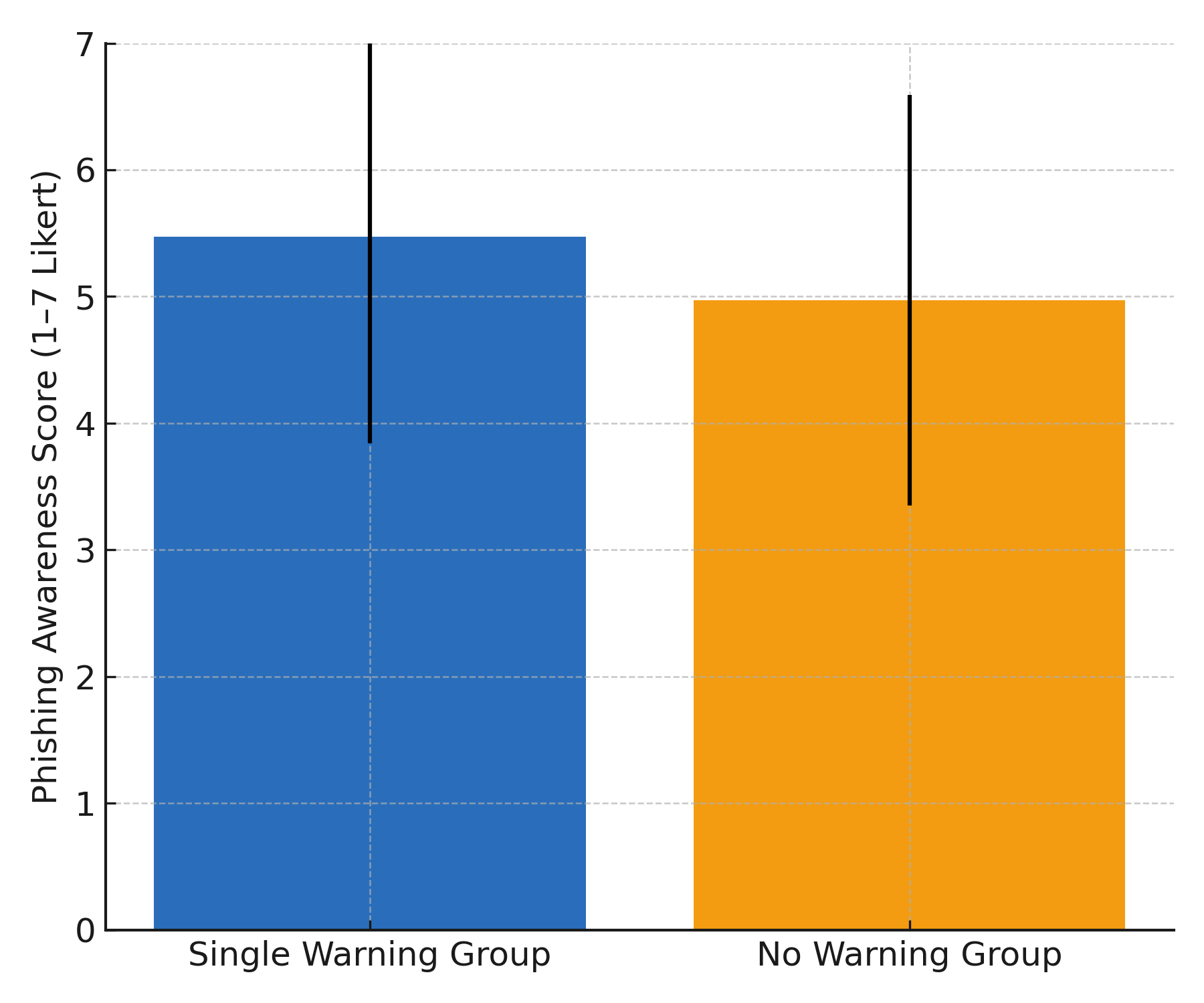}
\caption{Comparison of Scam Suspicion Score}\label{graph2}
\end{figure}

As shown in Table~\ref{tab:awareness_diff} and Figure~\ref{graph2}, the single-warning group reported slightly higher levels of suspicion compared to the control group. However, the difference was not statistically significant($t(58) = 0.96, p = .339$). The effect size was small (Cohen’s d = 0.25), and the 95\% confidence interval estimated through bootstrapping [–0.42, 1.22] also indicated a lack of statistical significance.

These results suggest that while a single warning may momentarily trigger suspicion, it is insufficient to sustain recipients’ psychological resistance throughout the full sequence of a strategically structured social engineering scam. In complex, dynamic threat scenarios such as voice phishing, more adaptive and context-aware interventions may be necessary to produce significant cognitive effects.

\begin{table}[h!]
\renewcommand{\arraystretch}{1.1}
\centering
\footnotesize
\begin{tabular}{p{0.8cm} p{0.05cm} p{0.3cm} p{0.3cm} p{0.3cm} p{0.5cm} p{1.4cm} p{0.6cm}}
\hline
\textbf{Group} & \textbf{N} & \textbf{M} & \textbf{SD} & \textbf{t(58)} & \textbf{P} & \textbf{95\% CI} & \tiny \textbf{Cohen's $d$} \\
\hline
\tiny Single Warning & 30 & 5.13 & 1.56 & 0.96 & .339 & [–0.42,1.22] & 0.25 \\
\tiny No Warning     & 30 & 4.73 & 1.66 &      &      &                 &      \\
\hline
\end{tabular}
\caption{Suspicion Scores of Single and No Warning. Cohen’s $d$ is reported as the standardized effect size.}
\label{tab:awareness_diff}
\end{table}

\subsection{RQ3: \textsc{ScriptMind} helps users recognize potential harm during scams} 
\label{RQ3}

\textbf{RQ3} aimed to evaluate whether a \textbf{\textsc{ScriptMind}} warning generated by LLMs could enhance users’ cognitive suspicion in response to a simulated SE attack. To examine this, we conducted a two-way repeated measures ANOVA, with five call stages as the within-subjects factor and experimental condition as the between-subjects factor.

As shown in Table~\ref{tab:anova_results}, the results revealed a significant main effect of stage on suspicion levels($F$(4, 348) = 23.79, $p$ $<$ .001, \text{partial } $\eta^2 = .215$). Importantly, a significant interaction effect between stage and group was also observed($F$(8, 348) = 2.15, $p$ = .031, $\text{partial } \eta^2 = .047$). This indicates that users’ suspicion did not merely fluctuate based on the temporal flow, but rather changed in distinct patterns depending on the type of warning.

\begin{table}[h!]
\renewcommand{\arraystretch}{1.1}
\centering
\small
\begin{tabular}{p{1.8cm} p{0.8cm} p{0.7cm} p{1cm} p{1.3cm}}
\hline
\textbf{Effect} & \textbf{df} & \textbf{F} & \textbf{P} & \textbf{Partial $\eta^2$} \\
\hline
Stage            & 4, 348   & 23.79 & $< .001$ & 0.215 \\
Stage × Group    & 8, 348   & 2.15  & .031     & 0.047 \\
\hline
\end{tabular}
\caption{Repeated Measures ANOVA Summary for Suspicion Scores by Stage and Each Group}
\label{tab:anova_results}
\end{table}

To better understand these patterns, we conducted stage-wise one-way ANOVAs comparing the groups at each of the stages. As shown in Table~\ref{tab:stage_group_stats}, the LLM Warning group exhibited the highest suspicion scores at Stage 4 and Stage 5, and the differences between groups at these stages were statistically significant ($p$ = .039, .024). No significant group differences were found in Stages 1-3.

\begin{table}[h!]
\renewcommand{\arraystretch}{1.2}
\centering
\tiny
\begin{tabular}{p{0.5cm} p{1.5cm} p{1.5cm} p{1cm} p{0.2cm} p{0.2cm}}
\hline
\textbf{Stage} & \textbf{LLM\_Warning} & \textbf{Single\_Warning} & \textbf{Control} & \textbf{F} & \textbf{P} \\
\hline
Stage1 & 5.63±1.69 & 5.40±1.92 & 5.07±2.07 & 0.67 & .512 \\
Stage2 & 5.60±1.65 & 4.77±2.08 & 4.80±1.90 & 1.88 & .159 \\
Stage3 & 4.63±1.99 & 3.87±2.21 & 4.13±1.96 & 1.07 & .346 \\
Stage4 & 6.27±1.60 & 6.00±1.49 & 5.23±1.72 & 3.36 & .039 \\
Stage5 & 5.73±2.05 & 5.63±1.69 & 4.43±2.25 & 3.88 & .024 \\
\hline
\end{tabular}
\caption{Suspicion Scores by Stage and Each Group}
\label{tab:stage_group_stats}
\end{table}

\begin{figure}[t]
\centering
\includegraphics[width=0.48\textwidth]{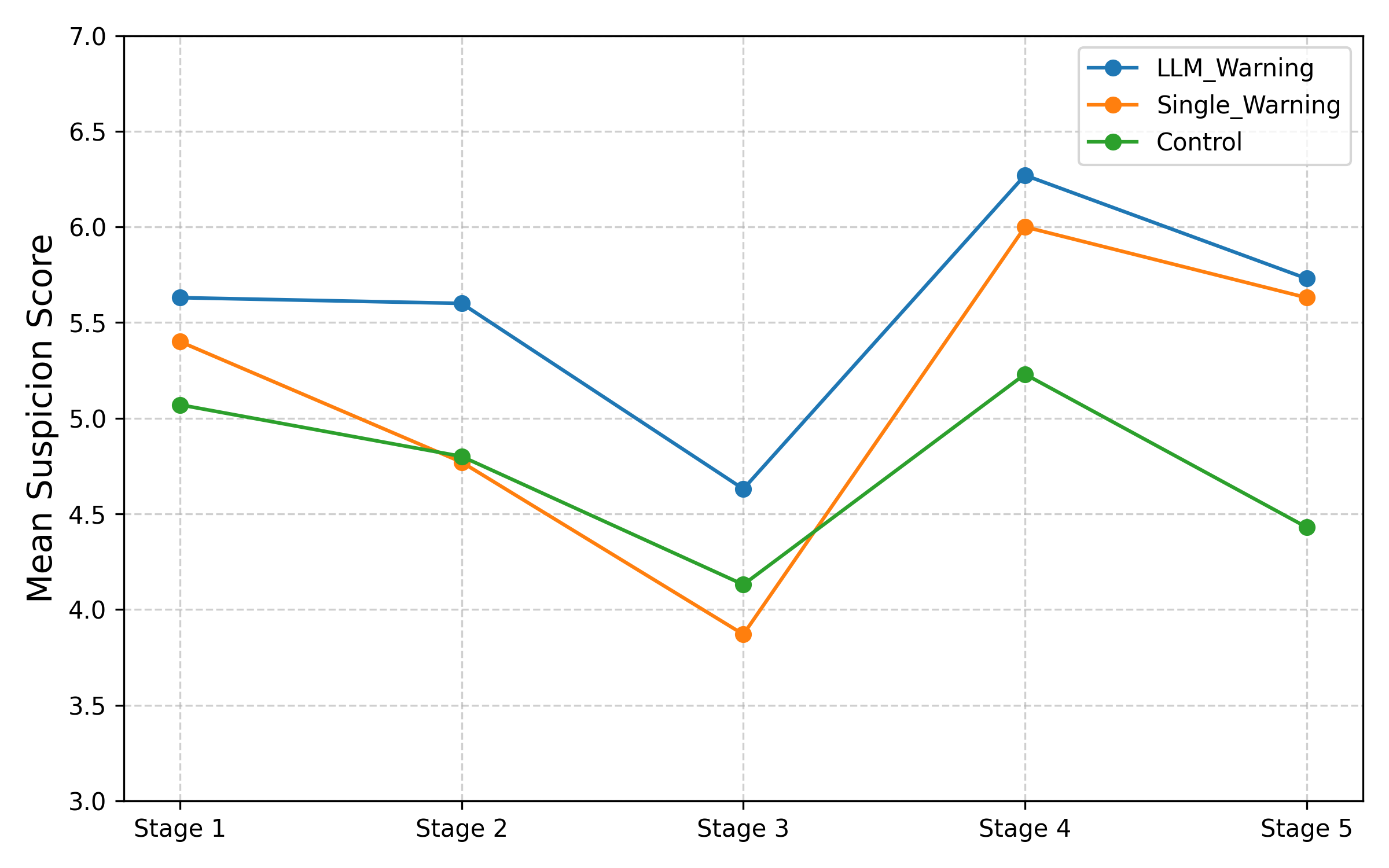}
\caption{Changes in Suspicion Levels by Script Stage}\label{graph3}
\end{figure}

Specifically, as visualized in Figure~\ref{graph3}, participants in the LLM Warning condition maintained relatively high levels of suspicion during the initial stages (Stages 1 and 2). Although their suspicion briefly declined at Stage 3, they showed a marked increase beginning at Stage 4, reaching the highest average suspicion levels by the final stage.

These findings suggest that \textbf{\textsc{ScriptMind}} interventions can elicit stronger and more sustained suspicion responses, especially as social engineering scam progresses into its critical stages involving pressure or financial requests. In contrast, participants in the control and single-warning groups exhibited either delayed or reduced recovery in suspicion following the mid-call drop in awareness.

In summary, the results support that LLM-based warnings are more effective than traditional single-message alerts in promoting cognitive resilience during phone scam. The dynamic and context-sensitive nature of \textbf{\textsc{ScriptMind}} predictions appears to better support users' psychological defense mechanisms, making this a promising intervention strategy for complex social engineering scams.

\end{document}